\pdfoutput=1

\documentclass[11pt]{article}

\usepackage[final]{acl2023}

\usepackage{times}
\usepackage{latexsym}
\usepackage{amsmath, amssymb}
\usepackage{xspace}
\usepackage{color, colortbl}
\usepackage{mathtools}
\usepackage{booktabs}
\usepackage{mdframed}
\usepackage[hang,flushmargin]{footmisc}
\usepackage{diagbox}
\usepackage{arydshln}
\usepackage{pifont}
\usepackage{paralist}

\newcommand{\miracl}[0]{MIRACL\xspace}
\newcommand{\nomiracl}[0]{NoMIRACL\xspace}

\newcommand\mrtydi{Mr.~\textsc{TyDi}\xspace}
\definecolor{paleaqua}{rgb}{0.74, 0.83, 0.9}

\usepackage{arydshln}

\definecolor{mygray}{gray}{0.2}
\usepackage[skins]{tcolorbox}
\newtcbox\tcbtp[1]{hbox, on line, colback=#1, enhanced, frame hidden, boxrule=0pt, 
    top=-1.5pt, bottom=-1.5pt, right=-2pt, left=-2pt, sharp corners}

\usepackage[T1]{fontenc}

\usepackage[utf8]{inputenc}

\usepackage{microtype}

\usepackage{inconsolata}

\title{``\emph{Knowing When You Don't Know}'': A Multilingual Relevance Assessment Dataset for Robust Retrieval-Augmented Generation}
\author{
Nandan Thakur$^1$, Luiz Bonifacio$^{1,3}$, Xinyu Zhang$^1$,
Odunayo Ogundepo$^1$, \\
\textbf{Ehsan Kamalloo}$^1$,
\textbf{David Alfonso-Hermelo}$^2$,
\textbf{Xiaoguang Li}$^2$,
\textbf{Qun Liu}$^2$, \\
\textbf{Boxing Chen}$^2$,
\textbf{Mehdi Rezagholizadeh}$^2$,
\textbf{Jimmy Lin}$^1$
\\[1ex]
$^1$ David R. Cheriton School of Computer Science, 
University of Waterloo, Canada \\
$^2$ Huawei Noah's Ark Lab \quad \quad $^3$ FEEC-Unicamp, Brazil
}

\begin{document}
\maketitle
\begin{abstract}
Retrieval-Augmented Generation (RAG) grounds Large Language Model (LLM) output by leveraging external knowledge sources to reduce factual hallucinations.
However, prior work lacks a comprehensive evaluation of different language families, making it challenging to evaluate LLM robustness against errors in external retrieved knowledge.
To overcome this, we establish~\textbf{\nomiracl}, a human-annotated dataset for evaluating LLM robustness in RAG across 18 typologically diverse languages.
\nomiracl includes both a non-relevant and a relevant subset. 
Queries in the non-relevant subset contain passages judged as non-relevant, whereas queries in the relevant subset include at least a single judged relevant passage. We measure relevance assessment using: 
(i) \textit{hallucination rate}, measuring model tendency to hallucinate, when the answer is not present in passages in the non-relevant subset, and (ii) \textit{error rate}, measuring model inaccuracy to recognize relevant passages in the relevant subset.
In our work, we observe that most models struggle to balance the two capacities. Models such as LLAMA-2 and Orca-2 achieve over 88\% hallucination rate on the non-relevant subset. Mistral and LLAMA-3 hallucinate less but can achieve up to a 74.9\% error rate on the relevant subset. Overall, GPT-4 is observed to provide the best tradeoff on both subsets, highlighting future work necessary to improve LLM robustness.
\nomiracl dataset and evaluation code are available at: \href{https://github.com/project-miracl/nomiracl}{https://github.com/project-miracl/nomiracl}.
\end{abstract}

\section{Introduction}

Retrieval-Augmented Generation (RAG) \citep{guu:2020, lewis:2020, izacard:2021, borgeaud:2022} is a promising way to incorporate external knowledge via a first-stage retrieval system.
RAG instills information from reliable knowledge corpora (provided as external passages) to generate accurate and faithful responses \cite{shuster:2021, gao:2023b, liu:2023}. 
Ever since the advent of Large Language Models (LLMs), such as GPT-3 \cite{brown:2020} or LLAMA-2 \cite{touvron:2023}, they are the de-facto choice for answer generation in RAG, due to their unprecedented progress in text generation and understanding \cite{brown:2020, li:2021, chang:2023, gao:2023c}.
RAG grounds the LLM-generated answer, thereby avoiding previously observed factual hallucination \cite{maynez:2020, raunak:2021} and outdated knowledge \cite{decao:2021, he:2023} in LLMs. 

\begin{figure}[t]
    \centering
    \begin{center}
        \includegraphics[trim=0 0 0 0,clip,width=0.47\textwidth]{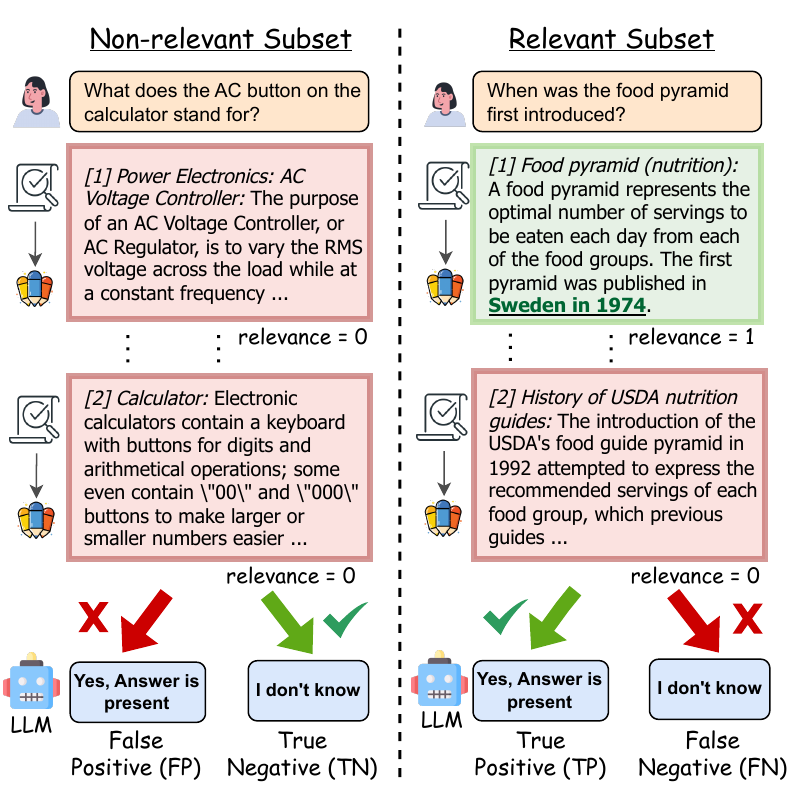}
        \caption{LLM robustness evaluation as a binary tree in \nomiracl. When dealing with queries in the non-relevant subset, the LLM is expected to disregard all noisy passages and refrain from answering ($\mathrm{TN}$). Conversely, for queries in the relevant subset, the LLM should recognize the relevant passage and provide a valid answer ($\mathrm{TP}$).}
        \label{fig:sap-vs-standard}
    \end{center}
    \vspace*{-\baselineskip}
\end{figure}

A challenging issue in RAG is to provide robust and reliable LLM-generated answers. 
The answer generation stage is dependent on the first-stage information retrieval system. The retrieval system poses challenges in accurately retrieving relevant information when evaluated either on zero-shot domains \cite{thakur:2021} or low-resource languages \cite{zhang:2023}. 
The incorrect or non-relevant information contained in retrieved passages can frequently mislead the LLM to hallucinate \cite{adlakha:2023, chen:2023, shi:2023, yoran:2023, yu:2023}. 
Prior work lacks a comprehensive evaluation of LLM reasoning capabilities in multiple languages.
As a result, it remains unclear to which extent LLMs hallucinate across both high- or low-resource languages.

To facilitate research in this direction, we present \textbf{\nomiracl}, a large multilingual human-annotated dataset containing over 56,000 (including both non-relevant and relevant samples) to evaluate LLM robustness against errors in first-stage external information, i.e., retrieved passages, across 18 typologically diverse languages. To construct the dataset,
we hired 31 native speakers as human annotators \cite{zhang:2023}.
\nomiracl contains two subsets, non-relevant and relevant. The non-relevant subset contains all queries with no known answers, i.e., all top-$k$ retrieved passages manually judged as non-relevant. Conversely, the relevant subset contains queries with known answers, i.e., at least one of the top-$k$ passages is manually judged as relevant. 

To better understand the LLM robustness in \nomiracl, we conduct experiments with several existing powerful and multilingual-focused LLMs (\emph{e.g.}, GPT-4, Mistral, LLAMA-3). We conduct our experiments using the top-$k$ oracle passages retrieved using a hybrid retrieval system from a language-specific Wikipedia corpus \cite{zhang:2023}. We use a zero-shot ``vanilla'' prompt template for prompting all LLMs. Our key findings are: First, LLMs such as LLAMA-2, Aya-101, and Orca-2 observe a surprisingly high 88\% hallucination rate on the non-relevant subset. Second, the Mistral and LLAMA-3 series of models hallucinate less but perform worse on the relevant subset. Overall, GPT-4 is found to provide the optimal performance tradeoff across both subsets.

To understand our experimental findings better, we conduct an empirical analysis on \nomiracl (\texttt{en}) to analyze the blind spots in a subset of LLMs. We observe LLAMA-2-7B and 13B interestingly repeat the query and prompt instructions on average by at least 25\%. Mistral-7B always provides a rationale in their output generation by over 88\%. In addition, we conduct different prompting techniques and observe that supervised fine-tuning LLMs on the \nomiracl development set can be tricky.

\begin{figure}[t]
    \centering
    \begin{center}
        \includegraphics[trim=0 5 0 0,clip,width=0.45\textwidth]{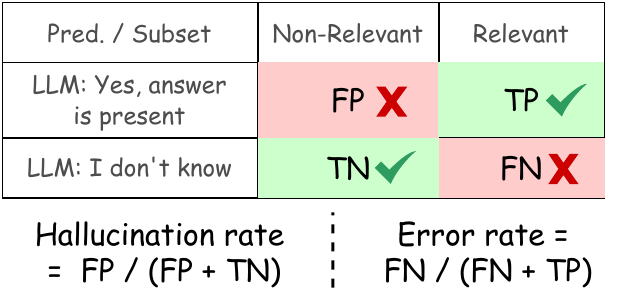}
        \caption{Confusion matrix for robustness evaluation with \nomiracl. More details are provided in (\S\ref{sec:robustness}); (Subset) denotes the ground-truth in \nomiracl; (Pred.) denotes the LLM output prediction.}
        \label{fig:contigency_table}
    \end{center}
    \vspace*{-\baselineskip}
\end{figure}

To summarize, our contributions are: (i) We introduce \nomiracl, a novel multilingual dataset to evaluate LLM hallucinations against first-stage retrieval errors in RAG. (ii) We evaluate several powerful multilingual LLMs and observe challenges in LLM robustness by often hallucinating an answer within non-answerable passages in the non-relevant subset and the inability to recognize relevant passages in the relevant subset. (iii) We conduct thorough manual inspections on LLM’s generation results, and find several hallucination patterns for each genre of the LLM; We hope \nomiracl can serve as a valuable dataset towards a much-needed LLM robustness evaluation.

\begin{figure*}[t!]
    \centering
    \begin{center}
        \includegraphics[trim=0 8 0 0,clip,width=\textwidth]{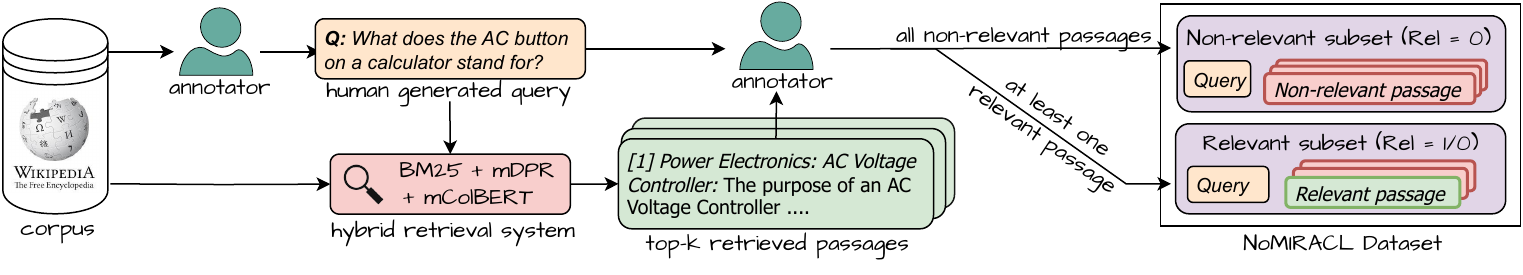}
        \caption{An overview of the data construction procedure (for English) involved in \nomiracl.}
        \label{fig:no-miracl-construction}
    \end{center}
    \vspace*{-\baselineskip}
\end{figure*}

\vspace{-2mm}
\section{Background and Problem Identification}
\vspace{-2mm}

A challenging issue in RAG is to provide robust and reliable LLM-generated output against a first-stage information retrieval system. Overreliance of the LLM on the content of the retrieved passages (i.e., tendency to extract information from passages) can be limiting when passages are noisy or non-relevant \cite{yu:2023, yoran:2023, shi:2023}.

\smallskip
\noindent\textbf{RAG Background.} Retrieval-augmented generation (RAG) \citep{lewis:2020, guu:2020} involves a two-stage inference pipeline. 
In the first stage, given the retrieval system and the user query, the retrieval system provides the subset of top-$k$ passages retrieved from an external data corpus $C$. For the next stage, the user query with the top-$k$ retrieved passages is provided to the LLM, which generates an output summarizing the answer for the query and citing the relevant passages.

\subsection{Robustness Evaluation}\label{sec:robustness}
We conduct our evaluation strategy as a contingency table (as shown in  \autoref{fig:contigency_table}) to robustly evaluate the LLM behavior in both answerable and non-answerable scenarios using a binary classification task, by comparing LLM predictions against the ground truth provided by human annotators.

\smallskip
\noindent\textbf{Definitions.} \nomiracl contains two subsets, we denote them as either non-relevant ($\mathrm{F}$) or relevant ($\mathrm{T}$). The non-relevant subset contains queries with no-known answers, i.e., all top-$k$ passages are non-relevant, while the relevant subset contains queries with known answers, i.e., at least one of the top-$k$ passages is relevant. Similarly, we denote the LLM prediction as either positive ($\mathrm{P}$) indicating the model finds the passage relevant to answer the query and similarly negative ($\mathrm{N}$) denotes the model does not find any passage relevant (i.e., containing the answer) for the query.

\smallskip
\noindent\textbf{Confusion Matrix.} In our confusion matrix (cf. \autoref{fig:contigency_table}), for our non-relevant subset, True Negative ($\mathrm{TN}$) denotes when the model correctly predicts queries with no-known answers using the non-relevant retrieved passages, whereas False Positive ($\mathrm{FP}$) denotes when the model prediction is incorrect on the non-relevant subset. 
Similarly, True Positive ($\mathrm{TP}$) denotes when LLM correctly predicts queries with known answers within the retrieved passages, whereas False Negative ($\mathrm{FN}$) denotes when the model prediction is incorrect (i.e., the model finds no answer) on the relevant subset.

\smallskip
\noindent\textbf{Evaluation Metrics.} Following prior works \cite{adlakha:2023, chen:2023}, we assess LLM robustness as a binary classification task using two metrics: (i) \emph{hallucination rate} and (ii) \emph{error rate}. First, we compute the hallucination rate (in \%) $=\mathrm{FP}/(\mathrm{FP} + \mathrm{TN})$ which measures the model's tendency to hallucinate an answer, when no answer is available in all of the passages in the non-relevant subset. Next, we measure the error rate (in \%) $=\mathrm{FN}/(\mathrm{FN} + \mathrm{TP})$ which measures the model's ability to identify the answer present within the passages in the relevant subset.

\begin{table*}[t]
\centering
\resizebox{\textwidth}{!}{
\begin{tabular}{l|ccccccccccccccccccc}
\toprule
 Split / ISO & \texttt{ar} & \texttt{bn} & \texttt{de} & \texttt{en} & \texttt{es} & \texttt{fa} & \texttt{fi} & \texttt{fr} & \texttt{hi} & \texttt{id} & \texttt{ja} & \texttt{ko} & \texttt{ru} & \texttt{sw} & \texttt{te} & \texttt{th} & \texttt{yo} & \texttt{zh} & Total \\
\specialrule{.4pt}{2pt}{0pt}
\rowcolor{paleaqua} \multicolumn{20}{l}{\textit{\textbf{Non-relevant Subset}: Queries with all human-judged non-relevant passages}} \\ 
Development & 228 & 495 & 171 & 289 & 245 & 760 & 98 & 1,016 & 1,016 & 474 & 211 & 1,577 & 268 & 508 & 480 & 323 & 1,678 & 1,085 & 10,922 \\
Test & 291 & 630 & 218 & 367 & 311 & 968 & 125 & 1,294 & 1294 & 603 & 269 & 2,006 & 342 & 646 & 610 & 412 & 2,136 & 1,381 & 13,903 \\
\specialrule{.4pt}{2pt}{0pt}
\rowcolor{paleaqua} \multicolumn{20}{l}{\textit{\textbf{Relevant Subset}: Queries with at least one human-judged relevant passage}} \\ 
Development & 2,896 & 411 & 305 & 799 & 648 & 632 & 1,271 & 343 & 350 & 960 & 860 & 213 & 1,252 & 482 & 828 & 733 & 119 & 393 & 13,495 \\
Test &  1,405 & 1,130 & 712 & 1,790 & 1,515 & 1,476 & 801 & 711 & 819 & 611 & 1,141 & 1,417 & 718 & 465 & 793 & 650 & 663 & 920 & 17,737 \\
\bottomrule
\end{tabular}
}
\caption{Dataset Statistics for \nomiracl. The dataset contains two subsets for all 18 languages: (i) Non-relevant subset, where queries contain all human-judged non-relevant passages. (ii) Relevant subset, where queries contain at least one relevant human-judged passage. Both subsets are split into disjoint development and test splits.
}
\label{tab:dataset-statistics}
\end{table*}

\section{\nomiracl Dataset}
As the goal of \nomiracl is to understand to which extent LLMs tend to hallucinate across different languages, our dataset contains 18 diverse languages with a myriad of both correct or answerable queries (relevant) subset and hallucinated or unanswerable queries (non-relevant). We describe our dataset construction procedure in (\S\ref{sec:data_construction}), fold creation in (\S\ref{sec:fold_creation}) and languages covered and dataset usage in (\S\ref{sec:languages_covered}). An overview of our data construction procedure is shown in \autoref{fig:no-miracl-construction}.

\subsection{Data Construction Procedure}\label{sec:data_construction}
\nomiracl is constructed using the same procedure utilized to develop MIRACL \cite{zhang:2023}. The data construction occurs in two stages, following \cite{zhang:2023, clark:2020}. In the first stage, the annotator (a native language speaker) writes a well-formed query for each individual prompt text. Each prompt is a short text snippet containing the first 100 words from a language-specific Wikipedia corpus. 
Next, for each human-generated query, top-$k$ passages are retrieved from the corpus using a hybrid multilingual retrieval system (more details in \S\ref{sec:eval_setup_and_metrics}).
In the second stage, annotators assess the binary relevance judgment of the top-$k$ query–passage pairs, either relevant (relevance = $1$) or non-relevant (relevance = $0$). For additional details in data construction, such as quality control, we would like to refer the reader to~\autoref{sec:additional_annotation_details}.

\smallskip\noindent\textbf{Non-relevant Subset.}
Annotators generate queries based on certain dataset guidelines, however, occasionally the human-generated queries cannot be answered with the external corpus, which leads to the scenario where none of the top-$k$ passages are relevant, i.e., none contains the answer. These queries with unknown answers may occur due to the following reasons: (i) queries can be either generic or specific for information to be present in Wikipedia, for e.g. ``\emph{What does the AC button on a calculator stand for?}'' retrieves the Wikipedia page on Calculator,\footnote{\href{https://en.wikipedia.org/wiki/Calculator}{https://en.wikipedia.org/wiki/Calculator}} but it does not contain information about the AC button; (ii) spelling mistakes in query generation. We construct the non-relevant subset with queries with all top-$k$ passages judged as non-relevant, i.e., with a relevance of $0$.

\smallskip\noindent\textbf{Relevant Subset.}
Queries with known answers, i.e. at least one of top-$k$ retrieved passages marked by the annotator as relevant to provide sufficient information to answer the query. We construct the \nomiracl relevant subset with queries with at least one relevant passage, i.e., all query-passage pairs have been judged as either relevant with a relevance of $1$ or non-relevant with relevance of $0$.

\subsection{Fold Creation}\label{sec:fold_creation}
In \nomiracl, we split the non-relevant and relevant subsets to form disjoint development and test splits.
Detailed statistics can be found in \autoref{tab:dataset-statistics}.

\smallskip
\noindent \textbf{Development Split.} For queries present in the relevant subset, we reuse the queries from the MIRACL development split \cite{zhang:2023} for all 18 languages. For queries in the non-relevant subset, we randomly sample a disjoint set containing 44\% of the queries from the whole non-relevant subset.

\smallskip
\noindent \textbf{Test Split.} For queries present in the relevant subset, we reuse the queries from the MIRACL test-B split \cite{zhang:2023} for all 18 languages.\footnote{We left out the MIRACL test-A split, as it contains queries for only 10 out of the 18 languages available.} For queries in the non-relevant subset, we utilize the other disjoint set, containing 56\% of the queries from the whole non-relevant subset.


\subsection{Languages Covered and Dataset Usage}\label{sec:languages_covered}
\nomiracl covers 18 diverse typological languages \cite{zhang:2023}.\footnote{\nomiracl covers 10 families (from Niger-Congo to Indo-European) and 11 scripts (from Latin to Devanagari) covering diversity from the perspective of linguistic characteristics.} The languages along with their ISO codes are: Arabic (\texttt{ar}), Bengali (\texttt{bn}), German (\texttt{de}), English (\texttt{en}), Spanish (\texttt{es}), Persian (\texttt{fa}), Finnish (\texttt{fi}), French (\texttt{fr}), Hindi (\texttt{hi}), Indonesian (\texttt{id}), Japanese (\texttt{ja}), Korean (\texttt{ko}), Russian (\texttt{ru}), Swahili (\texttt{sw}), Thai (\texttt{th}), Yoruba (\texttt{yo}), Chinese (\texttt{zh}). 

From \autoref{tab:dataset-statistics}, we observe an uneven amount of queries present in each language. To avoid this non-uniformity and budget constraints (see \autoref{sec:experimental_settings}), in our experiments, we limit the maximum number of 250 queries for each language and subset (if available)  in \nomiracl.

\section{Experimental Setup}

\subsection{Evaluation Setup and Metrics}\label{sec:eval_setup_and_metrics}
In \nomiracl, we assess LLM relevance as either hallucination or error, using an input query, a vanilla prompting technique, and top-$k$ (oracle) retrieved and relevance judged passages. 

\smallskip
\noindent\textbf{Retrieved Passages.} For each query in \nomiracl, a maximum of $k=10$ passages are retrieved and judged by our annotators. We follow the hybrid retrieval setup in \citet{zhang:2023}, which includes three different multilingual retriever models: (i) BM25 \cite{robertson:2009}, a lexical retriever, previously shown to be robust across domains and languages \cite{thakur:2021, zhang:2022}. We use the BM25 implementation available in Anserini \cite{yang:2018} with default parameters ($k_1=0.9$ and $b=0.4$) and the corresponding language-specific analyzer. (ii) mDPR \cite{karpukhin:2020}, a dense retriever, using mBERT \cite{devlin:2018} as the backbone and fine-tuned on MS MARCO with the Tevatron toolkit \cite{gao:2023}. (iii) mColBERT \cite{khattab:2020}, a multi-vector retriever, fine-tuned following \citet{khattab:2020} using mBERT as backbone and fine-tuned on MS MARCO. The top-$k$ passages are ranked using an ensemble fusion by normalizing and averaging each model score within the range of $[0, 1]$.

\smallskip
\noindent\textbf{Evaluation Objective.} In our work, we evaluate LLM relevance as a response string $y$ in a binary classification setup. Following prior evaluation strategies in \cite{adlakha:2023, yu:2023}, we use the input query $q_i$, a vanilla prompt template $Q$, and a set of top-$k$ annotated passages $P_k$. We prompt the LLM to evaluate if $q_i$ can be answered using any passage in $P_k$. The LLM generates an answer output containing either $y$ = ``\texttt{I don't know}'' as negative ($\mathrm{N}$) or
``\texttt{Yes, answer is present}'' as positive ($\mathrm{P}$). The output is tagged ``\texttt{Invalid}'' if it does not fall in either one of the above. Recall from (\S\ref{sec:robustness}),  we calculate the \emph{hallucination rate} (in \%) = $\mathrm{FP}/(\mathrm{FP} + \mathrm{TN})$, which measures the error in rejecting information from non-relevant passages and the \emph{error rate} (in \%) = $\mathrm{FN}/(\mathrm{FN} + \mathrm{TP}$), which measures the error in identifying relevant passages amongst noisy ones.

\subsection{Evaluation Models}
We evaluate eleven state-of-the-art LLMs with a strong focus of multilingual instruction capabilities, including both open and closed-sourced. All model checkpoints can be found in~\autoref{tab:model_links}. 

\smallskip
\noindent
\textbf{(1) OpenAI:} We include three closed-book LLM variants: GPT-3.5-turbo, GPT-4, and GPT-4o \cite{openai:2023} using the Azure OpenAI service. \textbf{(2) Mistral:} We include two variants: (i) \texttt{Mistral-7B-Instruct-v0.3}, the latest 7B instruction-tuned parameter model \cite{jiang:2023} and (ii) \texttt{Mixtral-8x7B-v0.1}, a sparse Mixture-of-Expert (MoE) model~\cite{jiang:2024}. \textbf{(3) Orca-2:} In the Orca-2 series \cite{mitra:2023}, we include both \texttt{Orca-2-7B} and \texttt{Orca-2-13B}.
\textbf{(4) Aya}: \texttt{Aya-101}~\cite{ustun2024aya} is a recently introduced multilingual LLM containing 13B parameters and trained with 101 languages and \texttt{Aya-23-35B} finetuned across 23 languages \cite{aryabumi:2024}. \textbf{(5) LLAMA-2:} In the LLAMA-2 series \cite{touvron:2023}, we include three chat variants: \texttt{Llama-2-7b-chat-hf}, \texttt{Llama-2-13b-chat-hf}, and \texttt{Llama-2-70b-chat-hf} instruction tuned chat models. \textbf{(6) LLAMA-3:} Following up on the LLAMA-2 series, we include both the instruction tuned models \cite{dubey:2024}: \texttt{Meta-Llama-3-8B-Instruct} and \texttt{Meta-Llama-3-70B-Instruct}.

\subsection{Experimental Settings}\label{sec:experimental_settings}
We execute the generation of GPT-4o, GPT-4, and GPT-3.5-turbo, using the OpenAI service (API version \texttt{2023-05-15}) deployed on Microsoft Azure\footnote{We compared Azure API with OpenAI API across four languages in \nomiracl and observed no noticeable difference between different GPT-4 API version providers.} and LLAMA-3 series using the AnyScale API service. We maintain a maximum input sequence length of 4096 tokens for a fair evaluation amongst all models. We set a low-temperature score = 0.1 for a deterministic output, and a top-p sampling ratio = 0.95. We output a maximum of $50$ tokens.

\begin{figure}[t!]
\begin{mdframed}[backgroundcolor=gray!5]
    \small
    \texttt{I will give you a question and several contexts containing information about the question. Read the contexts carefully. If any of the contexts answers the question, respond as either ``Yes, answer is present'' or ``I don't know'': \\\\
    QUESTION: \{query\} \\\\
    CONTEXTS:\\
    {[1]} \{Passage title\}: {\{Passage text\}}\\
    {[2]} \{Passage title\}: {\{Passage text\}}\\
    ... \\
    {[10]} \{Passage title\}: {\{Passage text\}}\\\\
    OUTPUT:
}
\end{mdframed}
\caption{Vanilla zero-shot prompt template used in our experiments for LLM hallucination evaluation for all 18 languages in \nomiracl. The instruction is provided in English, similar to \citet{ahuja:2023}.}
\label{fig:prompt_gpt4}
\end{figure}

\begin{figure*}[t]
    \centering
    \begin{center}
        \includegraphics[trim=0 0 0 0,clip,width=\textwidth]{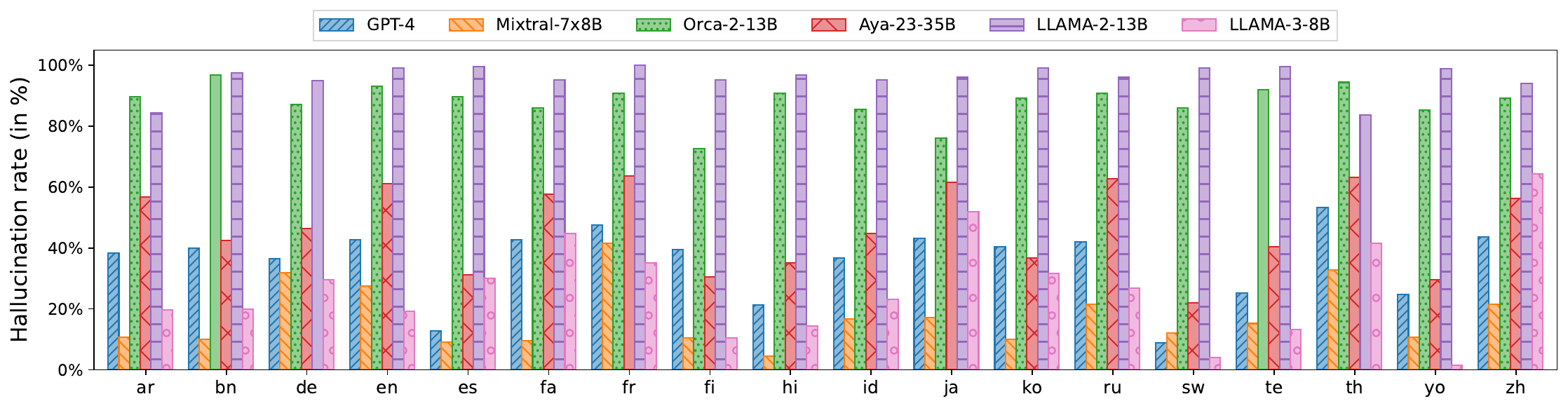}
        \caption{Hallucination rate (in \%) = $\mathrm{FP}/(\mathrm{FP} + \mathrm{TN}$) on the non-relevant subset ($\mathrm{F}$) in \nomiracl test split. The non-relevant subset contains queries with no known answers, i.e., all top-$k$ (where $k=10$) passages are judged by a human annotator as non-relevant. A majority of LLMs (except Mistral) hallucinate on the non-relevant subset. Lower the hallucination rate is better. The best model in each category is plotted (see \autoref{fig:non-relevant-baseline-results-all} for all models). }
        \label{fig:gpt-4-non-relevant-baseline-results}
    \end{center}
    \vspace*{-\baselineskip}
\end{figure*}

\begin{figure*}[t]
    \centering
    \begin{center}
        \includegraphics[trim=0 0 0 0,clip,width=\textwidth]{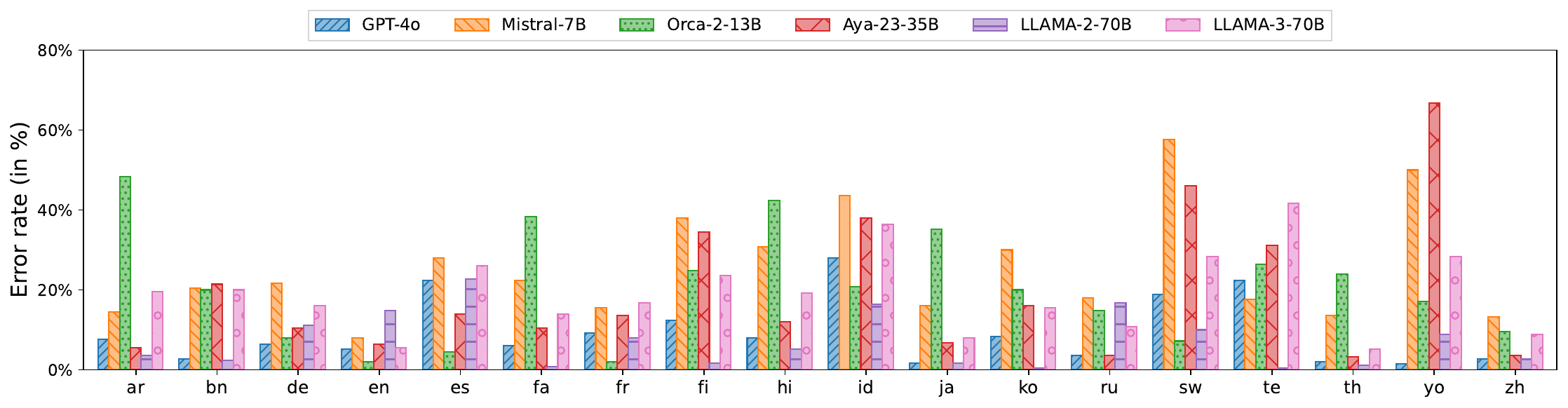}
        \caption{
        Error rate (in \%) = $\mathrm{FN}/(\mathrm{FN} + \mathrm{TP}$) on the relevant subset ($\mathrm{T}$) in \nomiracl test split. The relevant subset contains queries with known answers, i.e., at least one of the top-$k$ (where $k=10$) passages are judged by a human annotator as relevant. On average, a majority of LLMs (except Mistral and Aya-101) have a lower error rate by accurately identifying the answer. Lower the error rate is better. The best model in each category is plotted (see \autoref{fig:relevant-baseline-results-all} for all models).}
        \label{fig:gpt-4-relevant-baseline-results}
    \end{center}
    \vspace{-\baselineskip}
\end{figure*}

\smallskip
\noindent\textbf{Vanilla Prompting.} The choice of prompt significantly influences the performance and LLMs have been shown brittle to prompting variations, training examples, or long context setups \cite{liu:2023}. 
In our work, we evaluate all baselines using a \textit{zero-shot monolingual listwise} prompting strategy. We construct a vanilla prompt template using all top-$k$ (oracle) passages (available in \nomiracl) as a list of contexts along with the input query, both in the same language. We evaluate all LLMs zero-shot, as we cannot fit few-shot exemplars due to insufficient context length (maximum of 4096 sequence length). Our template provides a short description in English describing the task \citep{ahuja:2023}. Our vanilla prompt template used in our experiments is shown in \autoref{fig:prompt_gpt4}.

\smallskip
\noindent\textbf{Reducing Costs.} Running GPT-4 is expensive and LLAMA-3-70B is rather slow at inference. For long contexts and low-resource languages, the costs can even multiply. To limit, we did not exceed our prompt above 4096 tokens. To effectively fit all $k=10$ passages within the vanilla zero-shot prompt, we truncate each passage to use the first 375 tokens. Next, due to budget constraints, we keep our evaluation to a maximum of 250 randomly sampled queries for all languages in both \nomiracl relevant and non-relevant split. 
We end up providing $\approx$~20K API calls producing an expense of \$1,474 (in USD) including miscellaneous costs.

\vspace{-1mm}
\section{Experimental Results}
\vspace{-1mm}

We discuss our LLM robustness evaluation results using the hallucination rate on the non-relevant subset in (\S\ref{sec:non_relevant_results}) and using the error rate on the relevant subset in (\S\ref{sec:relevant_results}), and compare overall both the relevant and non-relevant capacities in (\S\ref{sec:overall_results}).

\subsection{\nomiracl Non-relevant Subset}\label{sec:non_relevant_results}
\autoref{fig:gpt-4-non-relevant-baseline-results} shows hallucination rates on the \nomiracl non-relevant subset for a maximum of 250 queries evaluated (each language) on all 18 languages for best LLM in each category (for all model results, please refer to \autoref{fig:non-relevant-baseline-results-all}). Our findings indicate that all LLMs (except Mistral) hallucinate that an answer is present across all languages, thereby indicating their poor ability to abstain from answering.
On average, the lowest hallucination rate of 17.4\% is observed by Mixtral-7x8B, followed by LLAMA-3-8B-Instruct with 26.8\%. 
GPT-4 achieves a 35.5\% hallucination rate, which highlights the challenge of identifying non-relevant passages. LLAMA-2, Orca-2, and Aya-101 perform much worse on average across all languages, by achieving a hallucination rate of more than 80\%. We hypothesize that LLMs perform poorly to identify non-relevant passages as they are highly similar to the query, but do not contain the exact answer.

Overall, the lowest hallucination rates are observed in Swahili and Yoruba. We hypothesize that queries in low-resource languages (smaller Wikipedia corpus) contain retrieved information, likely to be non-relevant (easier negative), thereby making it easier for the LLM to judge as ``I don't know''. GPT-3.5 (cf.~\autoref{fig:non-relevant-baseline-results-all}) is observed with the highest deviation across languages with a hallucination rate as low as 25.2\% on Swahili (\texttt{sw}) to 95.2\% on Bengali (\texttt{bn}).
Overall, all LLMs are found to perform poorly on \nomiracl non-relevant subset, indicating our dataset is very challenging in robustness evaluation for LLMs. 

\subsection{\nomiracl Relevant Subset}\label{sec:relevant_results}
\autoref{fig:gpt-4-relevant-baseline-results} shows error rates on \nomiracl relevant subset for a maximum of 250 queries (each language) on 18 languages for best LLM in each category 
Our findings indicate that all LLMs (except Mistral) identify the answer present within the relevant passage. 
On average, Aya-101 achieves the highest error rate of 62.5\%. The lowest error rates are observed by LLAMA-2-70B and GPT-4o which are lower than 10\%. Overall, Aya-101, Mixtral-7x8B, and LLAMA-3-8B perform worse on average by observing more than a 40\% error rate. Overall, LLMs (except Mistral and Aya-101) perform well and do not suffer from errors in identifying answers in the \nomiracl relevant subset. 

\subsection{\nomiracl Overall Comparison}\label{sec:overall_results}
\begin{figure}[t]
    \centering
    \begin{center}
        \includegraphics[trim=0 0 0 0,clip,width=0.48\textwidth]{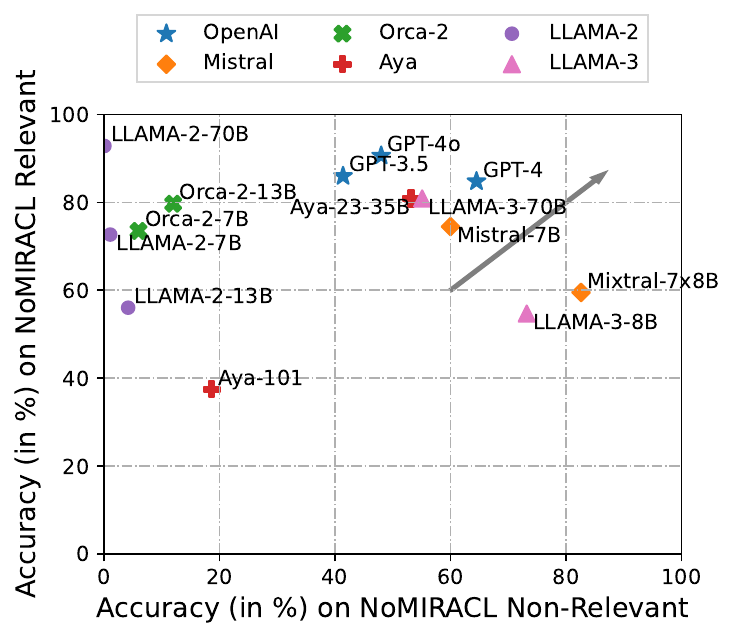}
        \caption{
        Plot measuring average model accuracy across all languages on relevant (y-axis) and non-relevant subset (x-axis) in \nomiracl. Performance towards the top-right corner (denoted by the arrow) is better.}
        \label{fig:average-model-results}
    \end{center}
    \vspace{-3mm}
\end{figure}

A robust LLM should be able to identify the answer captured within retrieved passages in the relevant subset and abstain from answering when none of the retrieved passages contain the answer in the non-relevant subset. 
To measure performance across both dimensions in \nomiracl, we plot the average model accuracy across both the non-relevant (x-axis) and relevant subset (y-axis) for all tested models in \autoref{fig:average-model-results}. Overall, LLMs positioned in the top-right corner provide an optimal performance on both subsets.
A majority of LLMs (such as LLAMA-2, and Orca-2) in the top-left corner perform well on the relevant subset, however, hallucinate and struggle to perform well on the non-relevant subset, indicating their inability to accurately judge non-relevant passages.

On the other hand, Mistral and LLAMA-3 suffer less from hallucination on the non-relevant subset but observe a higher error rate (over 40\%) on the relevant subset, indicating they are not confident in identifying passages containing the answer. Aya-101 is unable to perform well in either of the subsets. GPT-4 provides a good tradeoff balancing both a low hallucination and error rate on \nomiracl relevant and non-relevant subsets, however is expensive to compute at scale for inference.

\section{Empirical Analysis} 
In this section, we conduct an empirical analysis of LLM outputs, with both the non-relevant subset (containing hallucinations) and the relevant subset (containing errors). We conduct our ablation study on the English (\texttt{en}) subset in \nomiracl. We categorize each LLM output pattern as either positive or accurate (highlighted in \tcbtp{green!15}{green}), unable to understand instruction (highlighted in \tcbtp{orange!15}{orange}), or either a hallucination or error (highlighted in \tcbtp{red!15}{red}).

\smallskip
\noindent
\textbf{Non-relevant Subset.} As shown in \autoref{tab:empiricial_analysis_v2} (above), we observe a uniform distribution of the hallucination pattern of failed samples for a majority of LLMs by answering ``Yes, answer is present'' with or without additional explanation. LLMs such as LLAMA-2-7B and LLAMA-2-13B suffer from hallucinations by often repeating the question or instruction in their generation output. Mistral-7B interestingly always provides a rationale or explanation in their model response, whereas Aya-101, uses implicit memory heavily to directly provide an answer instead of grounding the answer from within the retrieved passages. Lastly, models such as Orca-2-7B tend to change the output generation style and often use synonyms such as ``No, answer is not present'' instead of ``I don't know''. 

\smallskip
\noindent
\textbf{Relevant Subset.} As shown in \autoref{tab:empiricial_analysis_v2} (below), similar to the non-relevant subset, we observe a uniform distribution of accurate LLM responses. Interestingly, GPT-4 and Orca-2-13B overall only provide a single output classification token, whereas models such as GPT-3.5 or Mixtral-8x7B provide an additional rationale or explanation. Similar to the non-relevant subset, LLAMA-2-7B and 13B models repeat the instruction in their output and Aya-101 sometimes uses implicit memory.

\begin{table*}[t!]
    \small
    \centering
    \resizebox{0.9\textwidth}{!}{\begin{tabular}{l r r r r r r r r r r r}
        \toprule
        & \multicolumn{2}{c}{\textbf{OpenAI}} & \multicolumn{2}{c}{\textbf{Mistral}} & \multicolumn{2}{c}{\textbf{Orca-2}} & \multicolumn{1}{c}{\textbf{Aya}} &  \multicolumn{3}{c}{\textbf{LLAMA-2}} \\
        \cmidrule(lr){2-3} \cmidrule(lr){4-5} \cmidrule(lr){6-7} \cmidrule(lr){8-8} \cmidrule(lr){9-11}
        & \textbf{GPT-4} & \textbf{GPT-3.5} & \textbf{8x7B} & \textbf{7B} & \textbf{13B} & \textbf{7B} & \textbf{Aya-101} & \textbf{70B} & \textbf{13B} & \textbf{7B} \\ \midrule
\multicolumn{11}{l}{\textit{\textbf{Empirical results on the non-relevant subset}: Queries with all human-judged non-relevant passages}} \\ \midrule
       \rowcolor{green!15} (i) Perfectly answers ``I don't know'' & 56.8\% & 54.8\% & 67.2\% & 1.6\% & 3.2\% & 9.2\% & 15.6\% & 0.0\% & 0.0\% & 0.8\% \\ 
        \rowcolor{green!15} (ii) ``I don't know'' with explanation & 0.0\% & 0.0\% & 3.6\% & 88.4\% & 0.0\% & 2.0\% & 0.0\% & 0.0\% & 0.0\% & 0.0\% \\
        \rowcolor{green!15} (iii) Uses a synonym of ``I don't know'' & 0.4\% & 0.0\% & 1.6\% & 1.2\% & 6.0\% & 7.6\% & 0.0\% & 0.4\% & 0.0\% & 2.4\% \\ \noalign{\vskip 0.5ex}\hdashline\noalign{\vskip 0.5ex}
        \rowcolor{orange!15} (iv) Refuses to answer & 0.4\% & 0.8\% & 0.0\% & 0.8\% & 3.2\% & 0.0\% & 6.4\% & 0.0\% & 0.0\% & 0.0\%  \\ 
        \rowcolor{orange!15} (v) Repeats question or instruction & 0.0\% & 0.0\% & 0.0\% & 0.0\% & 0.0\% & 3.2\% & 0.0\% & 2.0\% & 64.8\% & 30.8\% \\ 
        \rowcolor{orange!15} (vi) Conversation & 0.0\% & 0.0\% & 0.4\% & 0.0\% & 0.0\% & 0.4\% & 0.0\% & 0.0\% & 0.0\% & 0.4\% \\ \noalign{\vskip 0.5ex}\hdashline\noalign{\vskip 0.5ex}
        \rowcolor{red!15} (vii) Answers "Yes" w. or w.o. explanation & 42.4\% & 44.4\% & 27.2\% & 8.0\% & 87.6\% & 68.4\% & 67.2\% & 97.6\% & 34.8\% & 65.6\% \\ 
        \rowcolor{red!15} (viii) Uses implicit memory to answer & 0.0\% & 0.0\% & 0.0\% & 0.0\% & 0.0\% & 8.8\% & 10.8\% & 0.0\% & 0.0\% & 0.0\% \\ \midrule
        \multicolumn{11}{l}{\textit{\textbf{Empirical results on the relevant subset}: Queries with at least one human-judged relevant passage}} \\ \midrule
        \rowcolor{green!15} (i) Perfectly answers ``Yes, answer is present'' & 94.8\% & 45.2\% & 22.8\% & 5.2\% & 97.2\% &  71.2\% &  51.6\% & 1.2\% & 3.6\% & 7.2\% \\
        \rowcolor{green!15} (ii) "Yes, answer is present" with explanation & 0.0\% & 46.0\% & 56.0\% & 31.6\% & 0.8\% & 11.2\% & 0.0\% &  98.8\% &  46.8\% &  63.2\% \\ \noalign{\vskip 0.5ex}\hdashline\noalign{\vskip 0.5ex}
        \rowcolor{orange!15} (iii) Refuses to answer &  0.4\% &  0.8\% &  0.0\% & 0.0\% & 0.0\% & 0.0\% & 1.2\% & 0.0\% & 0.0\% & 0.0\%  \\
        \rowcolor{orange!15} (iv) Repeats question or instruction & 0.0\% & 0.0\% & 0.4\% & 0.0\% & 0.0\% & 0.4\% & 0.0\% & 0.0\% & 48.8\% & 24.8\% \\
        \rowcolor{orange!15} (v) Conversation & 0.0\% & 0.0\% & 0.0\% & 0.0\% & 0.0\% & 0.0\% & 0.0\% &  0.0\% & 0.0\% & 1.6\%\\ \noalign{\vskip 0.5ex}\hdashline\noalign{\vskip 0.5ex}
        \rowcolor{red!15} (vi) Answers "No" w. or w.o. explanation & 4.8\% &  8.0\% &  20.8\% & 60.8\% & 0.8\% & 2.8\% & 4.0\% &  0.0\% &  0.0\% &  1.2\%  \\
        \rowcolor{red!15} (vii) Uses implicit memory to answer & 0.0\% & 0.0\% & 0.0\% & 2.4\% & 1.2\% & 14.4\% & 42.0\% & 0.0\% &  0.0\% &  0.4\% \\
        \bottomrule
    \end{tabular}}
    \caption{Empirical results on the complete \nomiracl English (\texttt{en}) non-relevant (above) and relevant (below) subsets. The analysis is bracketed into three categories, where \tcbtp{green!15}{green} category denotes an accurate response, \tcbtp{orange!15}{orange} denotes limitations in understanding the instruction and \tcbtp{red!15}{red} denotes model hallucination or error respectively.
    \vspace{-\baselineskip}}
    \label{tab:empiricial_analysis_v2}
\end{table*}

\section{Further Studies}

\textbf{Prompt Optimization.} Prompting is crucial in handling the robustness evaluation of multilingual-focused LLMs. Techniques such as Chain-of-Thought (CoT) \cite{wei:2022} or algorithmically optimizing prompts using DSPy \cite{khattab:2024} highlight the necessity of prompt optimization. 
Although optimizing for the prompt is certainly challenging and expensive to evaluate all LLMs across 18 languages relevant and non-relevant subsets, we experiment with three listwise variations techniques inspired by \citet{thomas:2024}. The prompt template changes are listed in \autoref{fig:prompt_role_ablation}: (i) \textbf{role}, we highlight the role of LLM as an evaluator within the prompt at the beginning, (ii) \textbf{repeat}, we repeat the task instructions at the end of the prompt to remind the LLM, and (iii) \textbf{explanation}, we ask the LLM model to provide a step-by-step explanation and then answer and require 400 output tokens to fit both the LLM reasoning and the answer.

We evaluate Mistral-7B with three prompt variations independently on \nomiracl. The complete results are listed in \autoref{tab:prompt_optimization}. On average, both role and repeat techniques help reduce the error rate in the \nomiracl relevant subset by 6.3\% and 15.2\% but overall increase the hallucination rate by 8.7\% and 15.9\% respectively. On the other hand, prompting with explanation decreases the hallucination rate by 9.7\% but increases the error rate by 8.3\%. These results show that prompting is user dependent, the user will be required to choose their technique depending on whether they wish to be better on the non-relevant subset by reducing the hallucination rate or the relevant subset by reducing the error rate. 

\begin{table*}[t!]
    \centering
    \small
    \resizebox{\textwidth}{!}{\begin{tabular}{lccccccccccccccccccc}
        \toprule
        & \texttt{ar} & \texttt{bn} & \texttt{de} & \texttt{en} & \texttt{es} & \texttt{fa} & \texttt{fr} & \texttt{fi} & \texttt{hi} & \texttt{id} & \texttt{ja} & \texttt{ko} & \texttt{ru} & \texttt{sw} & \texttt{te} & \texttt{th} & \texttt{yo} & \texttt{zh} & \texttt{Avg.} \\
        \midrule
        \multicolumn{20}{l}{\emph{Hallucination Rates (in \%) on \nomiracl test split (non-relevant subset)}} \\ \midrule
        \textbf{Original} & 40.0 & 63.2 & 38.2 & 42.8 & 17.2 & 52.4 & 47.6 & 16.1 & 39.6 & 30.8 & 44.4 & 28.8 & 41.6 & 14.8 & 74.0 & 58.8 & 23.6 & 45.2 & 40.0 \\
        \textbf{\textcolor{blue}{(+ Role)}} & 35.6 & 60.0 & 53.5 & 62.4 & 40.4 & 38.0 & 71.2 & 29.8 & 38.4 & 49.2 & 59.2 & 41.2 & 55.6 & 21.2 & 72.0 & 54.4 & 32.4 & 62.8 & 48.7 \\
        \textbf{\textcolor{red}{(+ Repeat)}} & 47.2 & 72.4 & 56.7 & 50.8 & 35.2 & 69.2 & 70.0 & 50.0 & 49.6 & 48.8 & 65.2 & 48.4 & 56.0 & 35.2 & 78.4 & 72.0 & 43.6 & 58.4 & 55.9 \\
        \textbf{\textcolor{violet}{(+ Explanation)}} & 28.0 & 33.6 & 30.4 & 34.8 & 16.8 & 39.6 & 42.4 & 31.5 & 18.0 & 27.2 & 32.8 & 26.8 & 31.6 & 22.8 & 36.4 & 34.0 & 27.2 & 31.6 & 30.3 \\ \midrule
        \multicolumn{20}{l}{\emph{Error Rates (in \%) on \nomiracl test split (relevant subset)}} \\ \midrule
       \textbf{ Original} & 14.4 & 20.4 & 21.6 & 8.0 & 28.0 & 22.4 & 15.6 & 38.0 & 30.8 & 43.6 & 16.0 & 30.0 & 18.0 & 57.6 & 17.6 & 13.6 & 50.0 & 13.2 & 25.5 \\
        \textbf{\textcolor{blue}{(+ Role)}} & 15.2 & 23.6 & 10.0 & 3.2 & 14.4 & 23.2 & 7.2 & 29.2 & 25.2 & 32.8 & 9.6 & 21.6 & 10.8 & 47.2 & 17.6 & 17.2 & 33.8 & 4.4 & 19.2 \\
        \textbf{\textcolor{red}{(+ Repeat)}} & 8.4 & 12.4 & 9.2 & 5.2 & 12.4 & 4.8 & 5.6 & 6.0 & 16.4 & 16.8 & 1.6 & 12.0 & 6.8 & 22.8 & 14.0 & 6.0 & 21.6 & 3.6 & 10.3 \\
        \textbf{\textcolor{violet}{(+ Explanation)}} & 28.8 & 39.6 & 26.0 & 16.8 & 35.6 & 35.6 & 20.0 & 32.0 & 46.8 & 46.8 & 24.4 & 33.6 & 24.8 & 50.4 & 48.8 & 31.2 & 48.0 & 18.8 & 33.8 \\
        
        \bottomrule
    \end{tabular}}
    \caption{Hallucination and error rates on the \nomiracl test split (non-relevant and relevant subsets) with three types of prompting techniques on Mistral-7B (v0.3). The changes in the prompt template are listed in \autoref{fig:prompt_role_ablation}.}
    \label{tab:prompt_optimization}
\end{table*}

\smallskip
\noindent
\textbf{Fine-tuning on \nomiracl.} In this section, following prior works such as Chain-of-Verification \cite[CoVe;][]{dhuliawala:2023} or Chain-of-Noting \cite[CoN;][]{yu:2023}, we investigate the following research question: \emph{Does fine-tuning on the \nomiracl development set help increase robustness?} 

We experiment with two open-sourced LLMs: Mistral-7B and LLAMA-3 (8B). We Supervised Fine-Tune (SFT) LoRA adapters \cite{hu:2022} on the development set of \nomiracl for all 18 languages (randomly sampled 90\% train, 10\% development) using 4-A6000 GPUs each containing 48GB RAM with PEFT.\footnote{\href{https://github.com/huggingface/alignment-handbook}{https://github.com/huggingface/alignment-handbook}} Our hyperparameter settings are listed in \autoref{tab:hyperparameter_settings}. We were unable to fine-tune larger models (greater than 8B parameters) due to computational budget restrictions. 

As shown in \autoref{tab:fine-tuning-results}, we observe LLAMA-3 (8B) to be quite unstable after SFT. Fine-tuning helps to reduce the error rate of LLAMA-3 (8B) (an improvement of 10.6\%) but can hurt its performance on the hallucination rate (drop up to 17.9\%). For a few languages mentioned in \autoref{tab:fine-tuning-results-all} such as Arabic (ar) the LLM always outputs ``Yes, answer is present'', whereas for Bengali (bn) heavily relies on ``I don’t know''. On the other hand, SFT deteriorates Mistral-7B on both relevant and non-relevant datasets. Overall, we demonstrate SFT is tricky and careful experimentation is required to achieve the best out of fine-tuning on the \nomiracl development subset for a binary classification task output (``Yes, answer is present'' or ``I don’t know'').

\begin{table}[t!]
    \centering
    \small
    \begin{tabular}{lcl}
        \toprule
        \textbf{Model} & \textbf{w/o SFT} & \multicolumn{1}{c}{\textbf{w/ SFT}} \\
        \midrule
        \multicolumn{3}{l}{\emph{Non-Relevant Subset: Hallucination Rates (in \%)}} \\ \midrule
        Meta-Llama-3-8B-Instruct & \textbf{26.8} & 44.7 (-- 17.9) \\
        Mistral-Instruct-7B-v0.3 & \textbf{40.0} & 44.3 (-- 4.3) \\ \midrule
        \multicolumn{3}{l}{\emph{Relevant Subset: Error Rates (in \%)}} \\ \midrule
        Meta-Llama-3-8B-Instruct & 45.3 & \textbf{34.7} (+ 10.6) \\
        Mistral-Instruct-7B-v0.3 & \textbf{25.5} & 46.1 (-- 20.6) \\
        \bottomrule
    \end{tabular}
    \caption{Supervised fine-tuning on the \nomiracl development split with Llama-3 (8B) and Mistral-7B (v0.3) LLMs.}
    \label{tab:fine-tuning-results}
    \vspace{-\baselineskip}
\end{table}

\section{Related Work}

\textbf{Retrieval-Augmented Generation.} The knowledge stored in a large language model (LLM) is commonly outdated \cite{he:2023}, and prone to hallucinations by generating factually incorrect output \cite{maynez:2020, raunak:2021}.
By grounding on external knowledge, a retrieval-augmented LLM can generate better and more trustworthy output \cite{guu:2020, lewis:2020, izacard:2021, borgeaud:2022}.
Retrieval-augmented generation has achieved remarkable results in various tasks such as open-domain question answering (ODQA) \cite{lewis:2020, izacard:2021, trivedi:2023}, argument extraction \cite{du:2022} and code generation \cite{zhou:2023}.
Real-world products such as Bing Search and LangChain have incorporated RAG applications. 

\smallskip
\noindent\textbf{LLM Evaluation.} Prior work explores adding perturbation in passages and shows that LLM performance can be influenced when exposed to different tasks, such as question answering (QA) ~\cite{jia-liang-2017-adversarial, petroni:2020, creswell:2023}, logical reasoning ~\cite{misra:2023} or arithmetic reasoning ~\cite{shi:2023, kumar:2021}. 
In examining controllability and robustness,~\citet{li:2023} observes that LLMs disregard contextual information, showing that LLM output can be influenced by non-relevant context. ~\citet{adlakha:2023} observes complementary results from our work, where they observe LLM can be rather faithful when provided non-relevant passages in QA datasets such as NQ \cite{kwiatkowski:2019}.
Knowing that prompting LLMs with non-relevant data can result in misguided responses, \citet{yu:2023} recently introduced a new prompting technique, Chain-of-Noting (CON) and \citet{yoran:2023} fine-tuned the LLM explicitly, both aimed to improve LLM robustness in RAG when non-relevant information is provided.

\smallskip
\noindent\textbf{Related Datasets.} Datasets focused on addressing unanswerable queries such as SQuAD 2.0~\cite{rajpurkar-etal-2018-know} were created adversarially to look similar to datasets with answerable queries. Similarly, Conversational QA datasets such as CoQA~\cite{reddy:2019} and QuAC~\cite{choi-etal-2018-quac} also contain unanswerable queries. A concurrent work proposes RGB, a RAG benchmark to evaluate LLM robustness in English and Chinese \cite{chen:2023}. 

\section{Conclusion} 
We introduce \nomiracl, a multilingual human-labeled dataset for relevance assessment of LLM robustness as a binary relevance identification task in 18 languages. Our multilingual dataset is human-annotated and constructed with 31 native speakers.
We provide two subsets in \nomiracl, the non-relevant subset, where queries contain all judged non-relevant passages, and the relevant subset, where queries contain at least one relevant judged passage to measure the hallucination on the non-relevant and error on the relevant subset.
Our experimental results indicate that existing LLMs are not robust, as we observe challenges in LLM robustness in either hallucination or error. GPT-4 achieves the best model and performance tradeoff across both subsets. 
\nomiracl can facilitate research in understanding to which extent LLMs tend to hallucinate, ultimately paving the way for building more effective and robust multilingual-focused LLMs in the future.

\section{Limitations}
\nomiracl is not perfect and like other datasets have limitations. We describe our limitations below and keep it as future work to improve our dataset.

\smallskip
\noindent\textbf{1. Human Errors in Dataset Construction.} Our dataset has been fully constructed using humans, thereby it may contain human errors. We conducted additional quality checks on a subset of the \nomiracl dataset to validate its question quality and relevance judgment as explained in \autoref{sec:additional_annotation_details}.

\smallskip
\noindent\textbf{2. Evaluation Setup.} In our work, we evaluate whether a passage is relevant or non-relevant for a given query, instead of evaluating actual answer spans. Reliable and accurate answers for a given query require domain experts as annotators. Annotators can potentially highlight short extractive spans of answers within relevant passages, however, non-extractive queries can either contain multiple answers or a long-form answer, making it difficult to highlight a relevant answer span. Therefore, for \nomiracl, we focus on evaluating top-$k$ passages as information contexts, which are judged for their relevancy by a data annotator.

\smallskip
\noindent\textbf{3. Limited to Wikipedia.} \nomiracl is currently developed using language-specific Wikipedia as the corpora. Wikipedia may not be the ideal choice for real-world applications across languages. For example, the English BEIR benchmark \cite{thakur:2021} includes diversity within its domains (all English) and contains more real-world domains such as Medical, etc. However, we keep it as future work to extend \nomiracl to diverse domains for the following reasons: (i) \emph{scarcity of corpora across languages}: for low-resource languages such as Bengali or Yoruba, finding a suitable large enough text corpora is difficult with limited choices. (ii) \emph{no uniformity across domains}: certain European languages have more legal domain corpora available, whereas news articles for African languages. This will introduce non-uniformity in information across languages. (iii) \emph{limited budget}: constructing \nomiracl was expensive involving several annotators involved for about 4--6 months. Extending to more domains would require additional budgets and human effort to be able to implement.

\section*{Acknowledgements}
We would like to thank our annotators, without whom \nomiracl could not have been built. We would also like to thank Akintunde Oladipo for providing the necessary Microsoft Azure credits for evaluating OpenAI models. This research was supported in part by the Natural Sciences and Engineering Research Council (NSERC) of Canada, a gift from Huawei, and Cloud TPU support from Google’s TPU Research Cloud (TRC).

\bibliography{custom}

\begin{thebibliography}{58}
\expandafter\ifx\csname natexlab\endcsname\relax\def\natexlab#1{#1}\fi

\bibitem[{Adlakha et~al.(2024)Adlakha, BehnamGhader, Lu, Meade, and Reddy}]{adlakha:2023}
Vaibhav Adlakha, Parishad BehnamGhader, Xing~Han Lu, Nicholas Meade, and Siva Reddy. 2024.
\newblock \href {https://doi.org/10.1162/TACL\_A\_00667} {Evaluating correctness and faithfulness of instruction-following models for question answering}.
\newblock \emph{Trans. Assoc. Comput. Linguistics}, 12:681--699.

\bibitem[{Ahuja et~al.(2023)Ahuja, Diddee, Hada, Ochieng, Ramesh, Jain, Nambi, Ganu, Segal, Ahmed, Bali, and Sitaram}]{ahuja:2023}
Kabir Ahuja, Harshita Diddee, Rishav Hada, Millicent Ochieng, Krithika Ramesh, Prachi Jain, Akshay~Uttama Nambi, Tanuja Ganu, Sameer Segal, Mohamed Ahmed, Kalika Bali, and Sunayana Sitaram. 2023.
\newblock \href {https://doi.org/10.18653/V1/2023.EMNLP-MAIN.258} {{MEGA:} multilingual evaluation of generative {AI}}.
\newblock In \emph{Proceedings of the 2023 Conference on Empirical Methods in Natural Language Processing, {EMNLP} 2023, Singapore, December 6-10, 2023}, pages 4232--4267. Association for Computational Linguistics.

\bibitem[{Aryabumi et~al.(2024)Aryabumi, Dang, Talupuru, Dash, Cairuz, Lin, Venkitesh, Smith, Campos, Tan, Marchisio, Bartolo, Ruder, Locatelli, Kreutzer, Frosst, Gomez, Blunsom, Fadaee, {\"{U}}st{\"{u}}n, and Hooker}]{aryabumi:2024}
Viraat Aryabumi, John Dang, Dwarak Talupuru, Saurabh Dash, David Cairuz, Hangyu Lin, Bharat Venkitesh, Madeline Smith, Jon~Ander Campos, Yi~Chern Tan, Kelly Marchisio, Max Bartolo, Sebastian Ruder, Acyr Locatelli, Julia Kreutzer, Nick Frosst, Aidan~N. Gomez, Phil Blunsom, Marzieh Fadaee, Ahmet {\"{U}}st{\"{u}}n, and Sara Hooker. 2024.
\newblock \href {https://doi.org/10.48550/ARXIV.2405.15032} {Aya 23: Open weight releases to further multilingual progress}.
\newblock \emph{CoRR}, abs/2405.15032.

\bibitem[{Borgeaud et~al.(2022)Borgeaud, Mensch, Hoffmann, Cai, Rutherford, Millican, van~den Driessche, Lespiau, Damoc, Clark, de~Las~Casas, Guy, Menick, Ring, Hennigan, Huang, Maggiore, Jones, Cassirer, Brock, Paganini, Irving, Vinyals, Osindero, Simonyan, Rae, Elsen, and Sifre}]{borgeaud:2022}
Sebastian Borgeaud, Arthur Mensch, Jordan Hoffmann, Trevor Cai, Eliza Rutherford, Katie Millican, George van~den Driessche, Jean{-}Baptiste Lespiau, Bogdan Damoc, Aidan Clark, Diego de~Las~Casas, Aurelia Guy, Jacob Menick, Roman Ring, Tom Hennigan, Saffron Huang, Loren Maggiore, Chris Jones, Albin Cassirer, Andy Brock, Michela Paganini, Geoffrey Irving, Oriol Vinyals, Simon Osindero, Karen Simonyan, Jack~W. Rae, Erich Elsen, and Laurent Sifre. 2022.
\newblock \href {https://proceedings.mlr.press/v162/borgeaud22a.html} {Improving language models by retrieving from trillions of tokens}.
\newblock In \emph{International Conference on Machine Learning, {ICML} 2022, 17-23 July 2022, Baltimore, Maryland, {USA}}, volume 162 of \emph{Proceedings of Machine Learning Research}, pages 2206--2240. {PMLR}.

\bibitem[{Brown et~al.(2020)Brown, Mann, Ryder, Subbiah, Kaplan, Dhariwal, Neelakantan, Shyam, Sastry, Askell, Agarwal, Herbert{-}Voss, Krueger, Henighan, Child, Ramesh, Ziegler, Wu, Winter, Hesse, Chen, Sigler, Litwin, Gray, Chess, Clark, Berner, McCandlish, Radford, Sutskever, and Amodei}]{brown:2020}
Tom~B. Brown, Benjamin Mann, Nick Ryder, Melanie Subbiah, Jared Kaplan, Prafulla Dhariwal, Arvind Neelakantan, Pranav Shyam, Girish Sastry, Amanda Askell, Sandhini Agarwal, Ariel Herbert{-}Voss, Gretchen Krueger, Tom Henighan, Rewon Child, Aditya Ramesh, Daniel~M. Ziegler, Jeffrey Wu, Clemens Winter, Christopher Hesse, Mark Chen, Eric Sigler, Mateusz Litwin, Scott Gray, Benjamin Chess, Jack Clark, Christopher Berner, Sam McCandlish, Alec Radford, Ilya Sutskever, and Dario Amodei. 2020.
\newblock \href {https://proceedings.neurips.cc/paper/2020/hash/1457c0d6bfcb4967418bfb8ac142f64a-Abstract.html} {Language models are few-shot learners}.
\newblock In \emph{Advances in Neural Information Processing Systems 33: Annual Conference on Neural Information Processing Systems 2020, NeurIPS 2020, December 6-12, 2020, virtual}.

\bibitem[{Cao et~al.(2021)Cao, Aziz, and Titov}]{decao:2021}
Nicola~De Cao, Wilker Aziz, and Ivan Titov. 2021.
\newblock \href {https://doi.org/10.18653/V1/2021.EMNLP-MAIN.522} {Editing factual knowledge in language models}.
\newblock In \emph{Proceedings of the 2021 Conference on Empirical Methods in Natural Language Processing, {EMNLP} 2021, Virtual Event / Punta Cana, Dominican Republic, 7-11 November, 2021}, pages 6491--6506. Association for Computational Linguistics.

\bibitem[{Chang et~al.(2024)Chang, Wang, Wang, Wu, Yang, Zhu, Chen, Yi, Wang, Wang, Ye, Zhang, Chang, Yu, Yang, and Xie}]{chang:2023}
Yupeng Chang, Xu~Wang, Jindong Wang, Yuan Wu, Linyi Yang, Kaijie Zhu, Hao Chen, Xiaoyuan Yi, Cunxiang Wang, Yidong Wang, Wei Ye, Yue Zhang, Yi~Chang, Philip~S. Yu, Qiang Yang, and Xing Xie. 2024.
\newblock \href {https://doi.org/10.1145/3641289} {A survey on evaluation of large language models}.
\newblock \emph{{ACM} Trans. Intell. Syst. Technol.}, 15(3):39:1--39:45.

\bibitem[{Chen et~al.(2024)Chen, Lin, Han, and Sun}]{chen:2023}
Jiawei Chen, Hongyu Lin, Xianpei Han, and Le~Sun. 2024.
\newblock \href {https://doi.org/10.1609/AAAI.V38I16.29728} {Benchmarking large language models in retrieval-augmented generation}.
\newblock In \emph{Thirty-Eighth {AAAI} Conference on Artificial Intelligence, {AAAI} 2024, Thirty-Sixth Conference on Innovative Applications of Artificial Intelligence, {IAAI} 2024, Fourteenth Symposium on Educational Advances in Artificial Intelligence, {EAAI} 2014, February 20-27, 2024, Vancouver, Canada}, pages 17754--17762. {AAAI} Press.

\bibitem[{Choi et~al.(2018)Choi, He, Iyyer, Yatskar, Yih, Choi, Liang, and Zettlemoyer}]{choi-etal-2018-quac}
Eunsol Choi, He~He, Mohit Iyyer, Mark Yatskar, Wen-tau Yih, Yejin Choi, Percy Liang, and Luke Zettlemoyer. 2018.
\newblock \href {https://doi.org/10.18653/v1/D18-1241} {{Q}u{AC}: Question answering in context}.
\newblock In \emph{Proceedings of the 2018 Conference on Empirical Methods in Natural Language Processing}, pages 2174--2184, Brussels, Belgium. Association for Computational Linguistics.

\bibitem[{Clark et~al.(2020)Clark, Palomaki, Nikolaev, Choi, Garrette, Collins, and Kwiatkowski}]{clark:2020}
Jonathan~H. Clark, Jennimaria Palomaki, Vitaly Nikolaev, Eunsol Choi, Dan Garrette, Michael Collins, and Tom Kwiatkowski. 2020.
\newblock \href {https://doi.org/10.1162/TACL\_A\_00317} {Tydi {QA:} {A} benchmark for information-seeking question answering in typologically diverse languages}.
\newblock \emph{Trans. Assoc. Comput. Linguistics}, 8:454--470.

\bibitem[{Creswell et~al.(2023)Creswell, Shanahan, and Higgins}]{creswell:2023}
Antonia Creswell, Murray Shanahan, and Irina Higgins. 2023.
\newblock \href {https://openreview.net/pdf?id=3Pf3Wg6o-A4} {Selection-inference: Exploiting large language models for interpretable logical reasoning}.
\newblock In \emph{The Eleventh International Conference on Learning Representations, {ICLR} 2023, Kigali, Rwanda, May 1-5, 2023}. OpenReview.net.

\bibitem[{Devlin et~al.(2018)Devlin, Chang, Lee, and Toutanova}]{devlin:2018}
Jacob Devlin, Ming{-}Wei Chang, Kenton Lee, and Kristina Toutanova. 2018.
\newblock \href {http://arxiv.org/abs/1810.04805} {{BERT:} pre-training of deep bidirectional transformers for language understanding}.
\newblock \emph{CoRR}, abs/1810.04805.

\bibitem[{Dhuliawala et~al.(2024)Dhuliawala, Komeili, Xu, Raileanu, Li, Celikyilmaz, and Weston}]{dhuliawala:2023}
Shehzaad Dhuliawala, Mojtaba Komeili, Jing Xu, Roberta Raileanu, Xian Li, Asli Celikyilmaz, and Jason Weston. 2024.
\newblock \href {https://doi.org/10.18653/V1/2024.FINDINGS-ACL.212} {Chain-of-verification reduces hallucination in large language models}.
\newblock In \emph{Findings of the Association for Computational Linguistics, {ACL} 2024, Bangkok, Thailand and virtual meeting, August 11-16, 2024}, pages 3563--3578. Association for Computational Linguistics.

\bibitem[{Du and Ji(2022)}]{du:2022}
Xinya Du and Heng Ji. 2022.
\newblock \href {https://doi.org/10.18653/V1/2022.EMNLP-MAIN.307} {Retrieval-augmented generative question answering for event argument extraction}.
\newblock In \emph{Proceedings of the 2022 Conference on Empirical Methods in Natural Language Processing, {EMNLP} 2022, Abu Dhabi, United Arab Emirates, December 7-11, 2022}, pages 4649--4666. Association for Computational Linguistics.

\bibitem[{Dubey et~al.(2024)Dubey, Jauhri, Pandey, Kadian, Al{-}Dahle, Letman, Mathur, Schelten, Yang, Fan, Goyal, Hartshorn, Yang, Mitra, Sravankumar, Korenev, Hinsvark, Rao, Zhang, Rodriguez, Gregerson, Spataru, Rozi{\`{e}}re, Biron, Tang, Chern, Caucheteux, Nayak, Bi, Marra, McConnell, Keller, Touret, Wu, Wong, Ferrer, Nikolaidis, Allonsius, Song, Pintz, Livshits, Esiobu, Choudhary, Mahajan, Garcia{-}Olano, Perino, Hupkes, Lakomkin, AlBadawy, Lobanova, Dinan, Smith, Radenovic, Zhang, Synnaeve, Lee, Anderson, Nail, Mialon, Pang, Cucurell, Nguyen, Korevaar, Xu, Touvron, Zarov, Ibarra, Kloumann, Misra, Evtimov, Copet, Lee, Geffert, Vranes, Park, Mahadeokar, Shah, van~der Linde, Billock, Hong, Lee, Fu, Chi, Huang, Liu, Wang, Yu, Bitton, Spisak, Park, Rocca, Johnstun, Saxe, Jia, Alwala, Upasani, Plawiak, Li, Heafield, Stone, and et~al.}]{dubey:2024}
Abhimanyu Dubey, Abhinav Jauhri, Abhinav Pandey, Abhishek Kadian, Ahmad Al{-}Dahle, Aiesha Letman, Akhil Mathur, Alan Schelten, Amy Yang, Angela Fan, Anirudh Goyal, Anthony Hartshorn, Aobo Yang, Archi Mitra, Archie Sravankumar, Artem Korenev, Arthur Hinsvark, Arun Rao, Aston Zhang, Aur{\'{e}}lien Rodriguez, Austen Gregerson, Ava Spataru, Baptiste Rozi{\`{e}}re, Bethany Biron, Binh Tang, Bobbie Chern, Charlotte Caucheteux, Chaya Nayak, Chloe Bi, Chris Marra, Chris McConnell, Christian Keller, Christophe Touret, Chunyang Wu, Corinne Wong, Cristian~Canton Ferrer, Cyrus Nikolaidis, Damien Allonsius, Daniel Song, Danielle Pintz, Danny Livshits, David Esiobu, Dhruv Choudhary, Dhruv Mahajan, Diego Garcia{-}Olano, Diego Perino, Dieuwke Hupkes, Egor Lakomkin, Ehab AlBadawy, Elina Lobanova, Emily Dinan, Eric~Michael Smith, Filip Radenovic, Frank Zhang, Gabriel Synnaeve, Gabrielle Lee, Georgia~Lewis Anderson, Graeme Nail, Gr{\'{e}}goire Mialon, Guan Pang, Guillem Cucurell, Hailey Nguyen, Hannah Korevaar, Hu~Xu, Hugo
  Touvron, Iliyan Zarov, Imanol~Arrieta Ibarra, Isabel~M. Kloumann, Ishan Misra, Ivan Evtimov, Jade Copet, Jaewon Lee, Jan Geffert, Jana Vranes, Jason Park, Jay Mahadeokar, Jeet Shah, Jelmer van~der Linde, Jennifer Billock, Jenny Hong, Jenya Lee, Jeremy Fu, Jianfeng Chi, Jianyu Huang, Jiawen Liu, Jie Wang, Jiecao Yu, Joanna Bitton, Joe Spisak, Jongsoo Park, Joseph Rocca, Joshua Johnstun, Joshua Saxe, Junteng Jia, Kalyan~Vasuden Alwala, Kartikeya Upasani, Kate Plawiak, Ke~Li, Kenneth Heafield, Kevin Stone, and et~al. 2024.
\newblock \href {https://doi.org/10.48550/ARXIV.2407.21783} {The llama 3 herd of models}.
\newblock \emph{CoRR}, abs/2407.21783.

\bibitem[{Gao et~al.(2023{\natexlab{a}})Gao, Ma, Lin, and Callan}]{gao:2023}
Luyu Gao, Xueguang Ma, Jimmy Lin, and Jamie Callan. 2023{\natexlab{a}}.
\newblock \href {https://doi.org/10.1145/3539618.3591805} {Tevatron: An efficient and flexible toolkit for neural retrieval}.
\newblock In \emph{Proceedings of the 46th International {ACM} {SIGIR} Conference on Research and Development in Information Retrieval, {SIGIR} 2023, Taipei, Taiwan, July 23-27, 2023}, pages 3120--3124. {ACM}.

\bibitem[{Gao et~al.(2023{\natexlab{b}})Gao, Yen, Yu, and Chen}]{gao:2023b}
Tianyu Gao, Howard Yen, Jiatong Yu, and Danqi Chen. 2023{\natexlab{b}}.
\newblock \href {https://doi.org/10.18653/V1/2023.EMNLP-MAIN.398} {Enabling large language models to generate text with citations}.
\newblock In \emph{Proceedings of the 2023 Conference on Empirical Methods in Natural Language Processing, {EMNLP} 2023, Singapore, December 6-10, 2023}, pages 6465--6488. Association for Computational Linguistics.

\bibitem[{Guo et~al.(2023)Guo, Zhang, Wang, Jiang, Nie, Ding, Yue, and Wu}]{gao:2023c}
Biyang Guo, Xin Zhang, Ziyuan Wang, Minqi Jiang, Jinran Nie, Yuxuan Ding, Jianwei Yue, and Yupeng Wu. 2023.
\newblock \href {https://doi.org/10.48550/ARXIV.2301.07597} {How close is {C}hat{GPT} to human experts? comparison corpus, evaluation, and detection}.
\newblock \emph{CoRR}, abs/2301.07597.

\bibitem[{Guu et~al.(2020)Guu, Lee, Tung, Pasupat, and Chang}]{guu:2020}
Kelvin Guu, Kenton Lee, Zora Tung, Panupong Pasupat, and Ming{-}Wei Chang. 2020.
\newblock \href {http://proceedings.mlr.press/v119/guu20a.html} {Retrieval augmented language model pre-training}.
\newblock In \emph{Proceedings of the 37th International Conference on Machine Learning, {ICML} 2020, 13-18 July 2020, Virtual Event}, volume 119 of \emph{Proceedings of Machine Learning Research}, pages 3929--3938. {PMLR}.

\bibitem[{He et~al.(2023)He, Zhang, and Roth}]{he:2023}
Hangfeng He, Hongming Zhang, and Dan Roth. 2023.
\newblock \href {https://doi.org/10.48550/ARXIV.2301.00303} {Rethinking with retrieval: Faithful large language model inference}.
\newblock \emph{CoRR}, abs/2301.00303.

\bibitem[{Hu et~al.(2022)Hu, yelong shen, Wallis, Allen-Zhu, Li, Wang, Wang, and Chen}]{hu:2022}
Edward~J Hu, yelong shen, Phillip Wallis, Zeyuan Allen-Zhu, Yuanzhi Li, Shean Wang, Lu~Wang, and Weizhu Chen. 2022.
\newblock \href {https://openreview.net/forum?id=nZeVKeeFYf9} {Lo{RA}: Low-rank adaptation of large language models}.
\newblock In \emph{International Conference on Learning Representations}.

\bibitem[{Izacard and Grave(2021)}]{izacard:2021}
Gautier Izacard and Edouard Grave. 2021.
\newblock \href {https://doi.org/10.18653/V1/2021.EACL-MAIN.74} {Leveraging passage retrieval with generative models for open domain question answering}.
\newblock In \emph{Proceedings of the 16th Conference of the European Chapter of the Association for Computational Linguistics: Main Volume, {EACL} 2021, Online, April 19 - 23, 2021}, pages 874--880. Association for Computational Linguistics.

\bibitem[{Jia and Liang(2017)}]{jia-liang-2017-adversarial}
Robin Jia and Percy Liang. 2017.
\newblock \href {https://doi.org/10.18653/v1/D17-1215} {Adversarial examples for evaluating reading comprehension systems}.
\newblock In \emph{Proceedings of the 2017 Conference on Empirical Methods in Natural Language Processing}, pages 2021--2031, Copenhagen, Denmark. Association for Computational Linguistics.

\bibitem[{Jiang et~al.(2023)Jiang, Sablayrolles, Mensch, Bamford, Chaplot, de~Las~Casas, Bressand, Lengyel, Lample, Saulnier, Lavaud, Lachaux, Stock, Scao, Lavril, Wang, Lacroix, and Sayed}]{jiang:2023}
Albert~Q. Jiang, Alexandre Sablayrolles, Arthur Mensch, Chris Bamford, Devendra~Singh Chaplot, Diego de~Las~Casas, Florian Bressand, Gianna Lengyel, Guillaume Lample, Lucile Saulnier, L{\'{e}}lio~Renard Lavaud, Marie{-}Anne Lachaux, Pierre Stock, Teven~Le Scao, Thibaut Lavril, Thomas Wang, Timoth{\'{e}}e Lacroix, and William~El Sayed. 2023.
\newblock \href {https://doi.org/10.48550/ARXIV.2310.06825} {Mistral 7b}.
\newblock \emph{CoRR}, abs/2310.06825.

\bibitem[{Jiang et~al.(2024)Jiang, Sablayrolles, Roux, Mensch, Savary, Bamford, Chaplot, de~Las~Casas, Hanna, Bressand, Lengyel, Bour, Lample, Lavaud, Saulnier, Lachaux, Stock, Subramanian, Yang, Antoniak, Scao, Gervet, Lavril, Wang, Lacroix, and Sayed}]{jiang:2024}
Albert~Q. Jiang, Alexandre Sablayrolles, Antoine Roux, Arthur Mensch, Blanche Savary, Chris Bamford, Devendra~Singh Chaplot, Diego de~Las~Casas, Emma~Bou Hanna, Florian Bressand, Gianna Lengyel, Guillaume Bour, Guillaume Lample, L{\'{e}}lio~Renard Lavaud, Lucile Saulnier, Marie{-}Anne Lachaux, Pierre Stock, Sandeep Subramanian, Sophia Yang, Szymon Antoniak, Teven~Le Scao, Th{\'{e}}ophile Gervet, Thibaut Lavril, Thomas Wang, Timoth{\'{e}}e Lacroix, and William~El Sayed. 2024.
\newblock \href {https://doi.org/10.48550/ARXIV.2401.04088} {Mixtral of experts}.
\newblock \emph{CoRR}, abs/2401.04088.

\bibitem[{Karpukhin et~al.(2020)Karpukhin, Oguz, Min, Lewis, Wu, Edunov, Chen, and Yih}]{karpukhin:2020}
Vladimir Karpukhin, Barlas Oguz, Sewon Min, Patrick S.~H. Lewis, Ledell Wu, Sergey Edunov, Danqi Chen, and Wen{-}tau Yih. 2020.
\newblock \href {https://doi.org/10.18653/V1/2020.EMNLP-MAIN.550} {Dense passage retrieval for open-domain question answering}.
\newblock In \emph{Proceedings of the 2020 Conference on Empirical Methods in Natural Language Processing, {EMNLP} 2020, Online, November 16-20, 2020}, pages 6769--6781. Association for Computational Linguistics.

\bibitem[{Khattab et~al.(2023)Khattab, Singhvi, Maheshwari, Zhang, Santhanam, Vardhamanan, Haq, Sharma, Joshi, Moazam, Miller, Zaharia, and Potts}]{khattab:2024}
Omar Khattab, Arnav Singhvi, Paridhi Maheshwari, Zhiyuan Zhang, Keshav Santhanam, Sri Vardhamanan, Saiful Haq, Ashutosh Sharma, Thomas~T. Joshi, Hanna Moazam, Heather Miller, Matei Zaharia, and Christopher Potts. 2023.
\newblock \href {https://doi.org/10.48550/ARXIV.2310.03714} {{DSP}y: Compiling declarative language model calls into self-improving pipelines}.
\newblock \emph{CoRR}, abs/2310.03714.

\bibitem[{Khattab and Zaharia(2020)}]{khattab:2020}
Omar Khattab and Matei Zaharia. 2020.
\newblock \href {https://doi.org/10.1145/3397271.3401075} {Col{BERT}: Efficient and effective passage search via contextualized late interaction over {BERT}}.
\newblock In \emph{Proceedings of the 43rd International {ACM} {SIGIR} conference on research and development in Information Retrieval, {SIGIR} 2020, Virtual Event, China, July 25-30, 2020}, pages 39--48. {ACM}.

\bibitem[{Kumar et~al.(2021)Kumar, Maheshwary, and Pudi}]{kumar:2021}
Vivek Kumar, Rishabh Maheshwary, and Vikram Pudi. 2021.
\newblock \href {https://doi.org/10.18653/V1/2021.FINDINGS-EMNLP.230} {Adversarial examples for evaluating math word problem solvers}.
\newblock In \emph{Findings of the Association for Computational Linguistics: {EMNLP} 2021, Virtual Event / Punta Cana, Dominican Republic, 16-20 November, 2021}, pages 2705--2712. Association for Computational Linguistics.

\bibitem[{Kwiatkowski et~al.(2019)Kwiatkowski, Palomaki, Redfield, Collins, Parikh, Alberti, Epstein, Polosukhin, Devlin, Lee, Toutanova, Jones, Kelcey, Chang, Dai, Uszkoreit, Le, and Petrov}]{kwiatkowski:2019}
Tom Kwiatkowski, Jennimaria Palomaki, Olivia Redfield, Michael Collins, Ankur~P. Parikh, Chris Alberti, Danielle Epstein, Illia Polosukhin, Jacob Devlin, Kenton Lee, Kristina Toutanova, Llion Jones, Matthew Kelcey, Ming{-}Wei Chang, Andrew~M. Dai, Jakob Uszkoreit, Quoc Le, and Slav Petrov. 2019.
\newblock \href {https://doi.org/10.1162/TACL\_A\_00276} {Natural {Q}uestions: a benchmark for question answering research}.
\newblock \emph{Trans. Assoc. Comput. Linguistics}, 7:452--466.

\bibitem[{Lewis et~al.(2020)Lewis, Perez, Piktus, Petroni, Karpukhin, Goyal, K{\"{u}}ttler, Lewis, Yih, Rockt{\"{a}}schel, Riedel, and Kiela}]{lewis:2020}
Patrick S.~H. Lewis, Ethan Perez, Aleksandra Piktus, Fabio Petroni, Vladimir Karpukhin, Naman Goyal, Heinrich K{\"{u}}ttler, Mike Lewis, Wen{-}tau Yih, Tim Rockt{\"{a}}schel, Sebastian Riedel, and Douwe Kiela. 2020.
\newblock \href {https://proceedings.neurips.cc/paper/2020/hash/6b493230205f780e1bc26945df7481e5-Abstract.html} {Retrieval-augmented generation for knowledge-intensive {NLP} tasks}.
\newblock In \emph{Advances in Neural Information Processing Systems 33: Annual Conference on Neural Information Processing Systems 2020, NeurIPS 2020, December 6-12, 2020, virtual}.

\bibitem[{Li et~al.(2023)Li, Rawat, Zaheer, Wang, Lukasik, Veit, Yu, and Kumar}]{li:2023}
Daliang Li, Ankit~Singh Rawat, Manzil Zaheer, Xin Wang, Michal Lukasik, Andreas Veit, Felix~X. Yu, and Sanjiv Kumar. 2023.
\newblock \href {https://doi.org/10.18653/v1/2023.findings-acl.112} {Large language models with controllable working memory}.
\newblock In \emph{Findings of the Association for Computational Linguistics: {ACL} 2023, Toronto, Canada, July 9-14, 2023}, pages 1774--1793. Association for Computational Linguistics.

\bibitem[{Li et~al.(2024)Li, Tang, Zhao, Nie, and Wen}]{li:2021}
Junyi Li, Tianyi Tang, Wayne~Xin Zhao, Jian-Yun Nie, and Ji-Rong Wen. 2024.
\newblock \href {https://doi.org/10.1145/3649449} {Pre-trained language models for text generation: A survey}.
\newblock \emph{ACM Comput. Surv.}, 56(9).

\bibitem[{Liu et~al.(2024)Liu, Lin, Hewitt, Paranjape, Bevilacqua, Petroni, and Liang}]{liu:2023}
Nelson~F. Liu, Kevin Lin, John Hewitt, Ashwin Paranjape, Michele Bevilacqua, Fabio Petroni, and Percy Liang. 2024.
\newblock \href {https://doi.org/10.1162/TACL\_A\_00638} {Lost in the middle: How language models use long contexts}.
\newblock \emph{Trans. Assoc. Comput. Linguistics}, 12:157--173.

\bibitem[{Maynez et~al.(2020)Maynez, Narayan, Bohnet, and McDonald}]{maynez:2020}
Joshua Maynez, Shashi Narayan, Bernd Bohnet, and Ryan~T. McDonald. 2020.
\newblock \href {https://doi.org/10.18653/V1/2020.ACL-MAIN.173} {On faithfulness and factuality in abstractive summarization}.
\newblock In \emph{Proceedings of the 58th Annual Meeting of the Association for Computational Linguistics, {ACL} 2020, Online, July 5-10, 2020}, pages 1906--1919. Association for Computational Linguistics.

\bibitem[{Misra et~al.(2023)Misra, Rayz, and Ettinger}]{misra:2023}
Kanishka Misra, Julia Rayz, and Allyson Ettinger. 2023.
\newblock \href {https://doi.org/10.18653/v1/2023.eacl-main.213} {{COMPS}: Conceptual minimal pair sentences for testing robust property knowledge and its inheritance in pre-trained language models}.
\newblock In \emph{Proceedings of the 17th Conference of the European Chapter of the Association for Computational Linguistics}, pages 2928--2949, Dubrovnik, Croatia. Association for Computational Linguistics.

\bibitem[{Mitra et~al.(2023)Mitra, Corro, Mahajan, Codas, Sim{\~{o}}es, Agrawal, Chen, Razdaibiedina, Jones, Aggarwal, Palangi, Zheng, Rosset, Khanpour, and Awadallah}]{mitra:2023}
Arindam Mitra, Luciano~Del Corro, Shweti Mahajan, Andr{\'{e}}s Codas, Clarisse Sim{\~{o}}es, Sahaj Agrawal, Xuxi Chen, Anastasia Razdaibiedina, Erik Jones, Kriti Aggarwal, Hamid Palangi, Guoqing Zheng, Corby Rosset, Hamed Khanpour, and Ahmed Awadallah. 2023.
\newblock \href {https://doi.org/10.48550/ARXIV.2311.11045} {Orca 2: Teaching small language models how to reason}.
\newblock \emph{CoRR}, abs/2311.11045.

\bibitem[{OpenAI(2023)}]{openai:2023}
OpenAI. 2023.
\newblock \href {https://doi.org/10.48550/ARXIV.2303.08774} {{GPT-4} technical report}.
\newblock \emph{CoRR}, abs/2303.08774.

\bibitem[{Petroni et~al.(2020)Petroni, Lewis, Piktus, Rockt{\"{a}}schel, Wu, Miller, and Riedel}]{petroni:2020}
Fabio Petroni, Patrick S.~H. Lewis, Aleksandra Piktus, Tim Rockt{\"{a}}schel, Yuxiang Wu, Alexander~H. Miller, and Sebastian Riedel. 2020.
\newblock \href {https://doi.org/10.24432/C5201W} {How context affects language models' factual predictions}.
\newblock In \emph{Conference on Automated Knowledge Base Construction, {AKBC} 2020, Virtual, June 22-24, 2020}.

\bibitem[{Rajpurkar et~al.(2018)Rajpurkar, Jia, and Liang}]{rajpurkar-etal-2018-know}
Pranav Rajpurkar, Robin Jia, and Percy Liang. 2018.
\newblock \href {https://doi.org/10.18653/v1/P18-2124} {Know what you don{'}t know: Unanswerable questions for {SQ}u{AD}}.
\newblock In \emph{Proceedings of the 56th Annual Meeting of the Association for Computational Linguistics (Volume 2: Short Papers)}, pages 784--789, Melbourne, Australia. Association for Computational Linguistics.

\bibitem[{Raunak et~al.(2021)Raunak, Menezes, and Junczys{-}Dowmunt}]{raunak:2021}
Vikas Raunak, Arul Menezes, and Marcin Junczys{-}Dowmunt. 2021.
\newblock \href {https://doi.org/10.18653/V1/2021.NAACL-MAIN.92} {The curious case of hallucinations in neural machine translation}.
\newblock In \emph{Proceedings of the 2021 Conference of the North American Chapter of the Association for Computational Linguistics: Human Language Technologies, {NAACL-HLT} 2021, Online, June 6-11, 2021}, pages 1172--1183. Association for Computational Linguistics.

\bibitem[{Reddy et~al.(2019)Reddy, Chen, and Manning}]{reddy:2019}
Siva Reddy, Danqi Chen, and Christopher~D. Manning. 2019.
\newblock \href {https://doi.org/10.1162/tacl_a_00266} {{CoQA: A Conversational Question Answering Challenge}}.
\newblock \emph{Transactions of the Association for Computational Linguistics}, 7:249--266.

\bibitem[{Robertson and Zaragoza(2009)}]{robertson:2009}
Stephen~E. Robertson and Hugo Zaragoza. 2009.
\newblock \href {https://doi.org/10.1561/1500000019} {The probabilistic relevance framework: {BM25} and beyond}.
\newblock \emph{Found. Trends Inf. Retr.}, 3(4):333--389.

\bibitem[{Shi et~al.(2023)Shi, Chen, Misra, Scales, Dohan, Chi, Sch{\"{a}}rli, and Zhou}]{shi:2023}
Freda Shi, Xinyun Chen, Kanishka Misra, Nathan Scales, David Dohan, Ed~H. Chi, Nathanael Sch{\"{a}}rli, and Denny Zhou. 2023.
\newblock \href {https://proceedings.mlr.press/v202/shi23a.html} {Large language models can be easily distracted by irrelevant context}.
\newblock In \emph{International Conference on Machine Learning, {ICML} 2023, 23-29 July 2023, Honolulu, Hawaii, {USA}}, volume 202 of \emph{Proceedings of Machine Learning Research}, pages 31210--31227. {PMLR}.

\bibitem[{Shuster et~al.(2021)Shuster, Poff, Chen, Kiela, and Weston}]{shuster:2021}
Kurt Shuster, Spencer Poff, Moya Chen, Douwe Kiela, and Jason Weston. 2021.
\newblock \href {https://doi.org/10.18653/V1/2021.FINDINGS-EMNLP.320} {Retrieval augmentation reduces hallucination in conversation}.
\newblock In \emph{Findings of the Association for Computational Linguistics: {EMNLP} 2021, Virtual Event / Punta Cana, Dominican Republic, 16-20 November, 2021}, pages 3784--3803. Association for Computational Linguistics.

\bibitem[{Thakur et~al.(2021)Thakur, Reimers, R{\"{u}}ckl{\'{e}}, Srivastava, and Gurevych}]{thakur:2021}
Nandan Thakur, Nils Reimers, Andreas R{\"{u}}ckl{\'{e}}, Abhishek Srivastava, and Iryna Gurevych. 2021.
\newblock \href {https://datasets-benchmarks-proceedings.neurips.cc/paper/2021/hash/65b9eea6e1cc6bb9f0cd2a47751a186f-Abstract-round2.html} {{BEIR:} {A} heterogeneous benchmark for zero-shot evaluation of information retrieval models}.
\newblock In \emph{Proceedings of the Neural Information Processing Systems Track on Datasets and Benchmarks 1, NeurIPS Datasets and Benchmarks 2021, December 2021, virtual}.

\bibitem[{Thomas et~al.(2024)Thomas, Spielman, Craswell, and Mitra}]{thomas:2024}
Paul Thomas, Seth Spielman, Nick Craswell, and Bhaskar Mitra. 2024.
\newblock \href {https://doi.org/10.1145/3626772.3657707} {Large language models can accurately predict searcher preferences}.
\newblock In \emph{Proceedings of the 47th International {ACM} {SIGIR} Conference on Research and Development in Information Retrieval, {SIGIR} 2024, Washington DC, USA, July 14-18, 2024}, pages 1930--1940. {ACM}.

\bibitem[{Touvron et~al.(2023)Touvron, Martin, Stone, Albert, Almahairi, Babaei, Bashlykov, Batra, Bhargava, Bhosale, Bikel, Blecher, Canton{-}Ferrer, Chen, Cucurull, Esiobu, Fernandes, Fu, Fu, Fuller, Gao, Goswami, Goyal, Hartshorn, Hosseini, Hou, Inan, Kardas, Kerkez, Khabsa, Kloumann, Korenev, Koura, Lachaux, Lavril, Lee, Liskovich, Lu, Mao, Martinet, Mihaylov, Mishra, Molybog, Nie, Poulton, Reizenstein, Rungta, Saladi, Schelten, Silva, Smith, Subramanian, Tan, Tang, Taylor, Williams, Kuan, Xu, Yan, Zarov, Zhang, Fan, Kambadur, Narang, Rodriguez, Stojnic, Edunov, and Scialom}]{touvron:2023}
Hugo Touvron, Louis Martin, Kevin Stone, Peter Albert, Amjad Almahairi, Yasmine Babaei, Nikolay Bashlykov, Soumya Batra, Prajjwal Bhargava, Shruti Bhosale, Dan Bikel, Lukas Blecher, Cristian Canton{-}Ferrer, Moya Chen, Guillem Cucurull, David Esiobu, Jude Fernandes, Jeremy Fu, Wenyin Fu, Brian Fuller, Cynthia Gao, Vedanuj Goswami, Naman Goyal, Anthony Hartshorn, Saghar Hosseini, Rui Hou, Hakan Inan, Marcin Kardas, Viktor Kerkez, Madian Khabsa, Isabel Kloumann, Artem Korenev, Punit~Singh Koura, Marie{-}Anne Lachaux, Thibaut Lavril, Jenya Lee, Diana Liskovich, Yinghai Lu, Yuning Mao, Xavier Martinet, Todor Mihaylov, Pushkar Mishra, Igor Molybog, Yixin Nie, Andrew Poulton, Jeremy Reizenstein, Rashi Rungta, Kalyan Saladi, Alan Schelten, Ruan Silva, Eric~Michael Smith, Ranjan Subramanian, Xiaoqing~Ellen Tan, Binh Tang, Ross Taylor, Adina Williams, Jian~Xiang Kuan, Puxin Xu, Zheng Yan, Iliyan Zarov, Yuchen Zhang, Angela Fan, Melanie Kambadur, Sharan Narang, Aur{\'{e}}lien Rodriguez, Robert Stojnic, Sergey Edunov,
  and Thomas Scialom. 2023.
\newblock \href {https://doi.org/10.48550/ARXIV.2307.09288} {Llama 2: Open foundation and fine-tuned chat models}.
\newblock \emph{CoRR}, abs/2307.09288.

\bibitem[{Trivedi et~al.(2023)Trivedi, Balasubramanian, Khot, and Sabharwal}]{trivedi:2023}
Harsh Trivedi, Niranjan Balasubramanian, Tushar Khot, and Ashish Sabharwal. 2023.
\newblock \href {https://doi.org/10.18653/V1/2023.ACL-LONG.557} {Interleaving retrieval with chain-of-thought reasoning for knowledge-intensive multi-step questions}.
\newblock In \emph{Proceedings of the 61st Annual Meeting of the Association for Computational Linguistics (Volume 1: Long Papers), {ACL} 2023, Toronto, Canada, July 9-14, 2023}, pages 10014--10037. Association for Computational Linguistics.

\bibitem[{{\"{U}}st{\"{u}}n et~al.(2024){\"{U}}st{\"{u}}n, Aryabumi, Yong, Ko, D'souza, Onilude, Bhandari, Singh, Ooi, Kayid, Vargus, Blunsom, Longpre, Muennighoff, Fadaee, Kreutzer, and Hooker}]{ustun2024aya}
Ahmet {\"{U}}st{\"{u}}n, Viraat Aryabumi, Zheng~Xin Yong, Wei{-}Yin Ko, Daniel D'souza, Gbemileke Onilude, Neel Bhandari, Shivalika Singh, Hui{-}Lee Ooi, Amr Kayid, Freddie Vargus, Phil Blunsom, Shayne Longpre, Niklas Muennighoff, Marzieh Fadaee, Julia Kreutzer, and Sara Hooker. 2024.
\newblock \href {https://doi.org/10.18653/V1/2024.ACL-LONG.845} {Aya model: An instruction finetuned open-access multilingual language model}.
\newblock In \emph{Proceedings of the 62nd Annual Meeting of the Association for Computational Linguistics (Volume 1: Long Papers), {ACL} 2024, Bangkok, Thailand, August 11-16, 2024}, pages 15894--15939. Association for Computational Linguistics.

\bibitem[{Voorhees(1998)}]{voorhees:1998}
Ellen~M. Voorhees. 1998.
\newblock \href {https://doi.org/10.1145/290941.291017} {Variations in relevance judgments and the measurement of retrieval effectiveness}.
\newblock In \emph{{SIGIR} '98: Proceedings of the 21st Annual International {ACM} {SIGIR} Conference on Research and Development in Information Retrieval, August 24-28 1998, Melbourne, Australia}, pages 315--323. {ACM}.

\bibitem[{Wei et~al.(2022)Wei, Wang, Schuurmans, Bosma, Ichter, Xia, Chi, Le, and Zhou}]{wei:2022}
Jason Wei, Xuezhi Wang, Dale Schuurmans, Maarten Bosma, Brian Ichter, Fei Xia, Ed~H. Chi, Quoc~V. Le, and Denny Zhou. 2022.
\newblock \href {http://papers.nips.cc/paper\_files/paper/2022/hash/9d5609613524ecf4f15af0f7b31abca4-Abstract-Conference.html} {Chain-of-thought prompting elicits reasoning in large language models}.
\newblock In \emph{NeurIPS}.

\bibitem[{Yang et~al.(2018)Yang, Fang, and Lin}]{yang:2018}
Peilin Yang, Hui Fang, and Jimmy Lin. 2018.
\newblock \href {https://doi.org/10.1145/3239571} {Anserini: Reproducible ranking baselines using {L}ucene}.
\newblock \emph{{ACM} J. Data Inf. Qual.}, 10(4):16:1--16:20.

\bibitem[{Yoran et~al.(2024)Yoran, Wolfson, Ram, and Berant}]{yoran:2023}
Ori Yoran, Tomer Wolfson, Ori Ram, and Jonathan Berant. 2024.
\newblock \href {https://openreview.net/forum?id=ZS4m74kZpH} {Making retrieval-augmented language models robust to irrelevant context}.
\newblock In \emph{The Twelfth International Conference on Learning Representations, {ICLR} 2024, Vienna, Austria, May 7-11, 2024}. OpenReview.net.

\bibitem[{Yu et~al.(2023)Yu, Zhang, Pan, Ma, Wang, and Yu}]{yu:2023}
Wenhao Yu, Hongming Zhang, Xiaoman Pan, Kaixin Ma, Hongwei Wang, and Dong Yu. 2023.
\newblock \href {https://doi.org/10.48550/ARXIV.2311.09210} {Chain-of-note: Enhancing robustness in retrieval-augmented language models}.
\newblock \emph{CoRR}, abs/2311.09210.

\bibitem[{Zhang et~al.(2022)Zhang, Ogueji, Ma, and Lin}]{zhang:2022}
Xinyu Zhang, Kelechi Ogueji, Xueguang Ma, and Jimmy Lin. 2022.
\newblock \href {https://doi.org/10.48550/ARXIV.2204.02363} {Towards best practices for training multilingual dense retrieval models}.
\newblock \emph{CoRR}, abs/2204.02363.

\bibitem[{Zhang et~al.(2023)Zhang, Thakur, Ogundepo, Kamalloo, Alfonso-Hermelo, Li, Liu, Rezagholizadeh, and Lin}]{zhang:2023}
Xinyu Zhang, Nandan Thakur, Odunayo Ogundepo, Ehsan Kamalloo, David Alfonso-Hermelo, Xiaoguang Li, Qun Liu, Mehdi Rezagholizadeh, and Jimmy Lin. 2023.
\newblock \href {https://doi.org/10.1162/tacl_a_00595} {{{MIRACL}: A Multilingual Retrieval Dataset Covering 18 Diverse Languages}}.
\newblock \emph{Transactions of the Association for Computational Linguistics}, 11:1114--1131.

\bibitem[{Zhou et~al.(2023)Zhou, Alon, Xu, Jiang, and Neubig}]{zhou:2023}
Shuyan Zhou, Uri Alon, Frank~F. Xu, Zhengbao Jiang, and Graham Neubig. 2023.
\newblock \href {https://openreview.net/pdf?id=ZTCxT2t2Ru} {Doc{P}rompting: Generating code by retrieving the docs}.
\newblock In \emph{The Eleventh International Conference on Learning Representations, {ICLR} 2023, Kigali, Rwanda, May 1-5, 2023}. OpenReview.net.

\end{thebibliography}

\clearpage

\appendix

\section{Appendix}
\label{sec:appendix}
The following supplementary sections in the appendix are arranged as follows:
\begin{compactitem}
    \item \autoref{sec:dataset-release} provides information on the \nomiracl dataset release.
    \item \autoref{sec:additional_annotation_details} provides additional construction details in \nomiracl, including corpora preparation and annotator hiring details.
    \item \autoref{sec:quality_control} describes steps we took for quality control during the dataset construction.
    \item \autoref{sec:additional_details} provides model checkpoints and additional experimental results.
\end{compactitem}

\section{Details on \nomiracl Dataset Release}\label{sec:dataset-release}

\smallskip
\noindent\textbf{Licensing.} The \nomiracl dataset is based on language-specific Wikipedia. We follow the same license as Wikipedia for \nomiracl: Creative Commons Attribution-ShareAlike 4.0 Unported License (CC BY-SA 4.0).\footnote{\url{https://creativecommons.org/licenses/by-sa/4.0/}} Overall, the license allows both
researchers and industry alike to access the dataset, and allow them to copy and redistribute the dataset for future work.

\smallskip
\noindent\textbf{Examples.} A randomly sampled example for each of the non-relevant and relevant subsets of the \nomiracl dataset for English (\texttt{en}) has been provided in \autoref{tab:examples_non_valid} and \autoref{tab:examples_valid} respectively.

\section{Additional Data Construction Details}\label{sec:additional_annotation_details}

\noindent\textbf{Corpora Preparation.} For each \nomiracl language, we follow the same passage corpora provided in \miracl~\cite{zhang:2023}. Out of the 18 languages, 11 of the existing languages common in \mrtydi~\cite{zhang:2022} use the raw Wikipedia dump from early 2019 and the rest of the languages used in \miracl use a release from March 2022. In \miracl, all Wikipedia articles are parsed using WikiExtractor\footnote{\url{https://github.com/attardi/wikiextractor}}
and segmented into passages based on natural discourse units using two consecutive newlines in the wiki markup as the delimiter.

\smallskip
\noindent\textbf{Annotator Hiring Details.} An important feature of \nomiracl is that our dataset was {\it not} constructed via crowd-sourced workers similar to \miracl~\cite{zhang:2023}. We overall hired 31 annotators (both part-time and full-time) across all languages in \nomiracl.
Each annotator was interviewed and evaluated to be a native speaker of their language, based on a carefully constructed onboarding and training process. Overall our hiring process and \nomiracl data construction in total took around 6 months. We offered annotators the hourly rate of \$18.50 per hour (converted into USD). For reference, the local minimum wage is \$11.50 USD/hr.

\section{Quality Control}\label{sec:quality_control}
To ensure high data quality, we conduct a manual assessment executed by human reviewers (hired part-time) on a random subset of \nomiracl annotations, following \miracl~\cite{zhang:2023}. We conducted our quality control in two phases.

\smallskip
\noindent{\textbf{Phase I.}}
In this phase, reviewers were given both the prompts and the generated queries and filled up a checklist to determine whether the quality of the queries met our requirements. Criteria include the examination of the query itself (e.g., spelling, syntax, and fluency, etc.) and whether the query could be answered directly by the prompt, which we wanted to avoid to generate more informative queries, following \cite{clark:2020, zhang:2023}. To evaluate this, we measured the lexical overlap between the queries and their corresponding prompts. We found the overlaps primarily occur in entities or stopwords and thus concluded that the generated queries are reasonably different from the given prompts.

\smallskip
\noindent{\textbf{Phase II.}}
In this phase, reviewers were provided the same guidance as annotators performing the relevance assessment. They were asked to label a randomly sampled subset of the query--passage pairs from our annotated batch.
The degree of agreement on the overlapping pairs is used to quantify the quality of the relevance labels. Overall, we observe on average agreements of over 80\% on query--passage relevance, which is consistent with the IR literature dating back many decades~\cite{voorhees:1998}.

\section{Checkpoints and Additional Results}\label{sec:additional_details}
\begin{figure*}[tb]
    \centering
    \begin{center}
        \includegraphics[trim=0 0 0 0,clip,width=\textwidth]{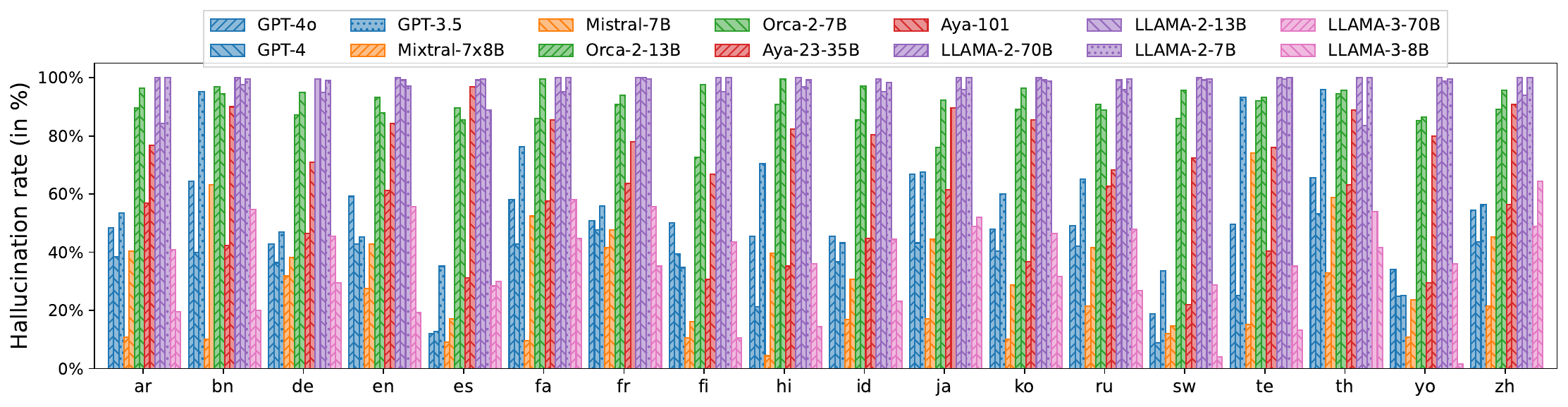}
        \caption{Hallucination rate (in \%) = $\mathrm{FP}/(\mathrm{FP} + \mathrm{TN}$) on the non-relevant subset in \nomiracl test split. The non-relevant subset contains queries with no-known answers, i.e., all top-$k$ (where $k=10$) passages are judged by a human annotator as non-relevant. On average, most LLMs (except Mistral) hallucinate on the non-relevant subset. Lower the hallucination rate is better.}
        \label{fig:non-relevant-baseline-results-all}
    \end{center}
    \vspace*{-\baselineskip}
\end{figure*}

\begin{figure*}[tb]
    \centering
    \begin{center}
        \includegraphics[trim=0 0 0 0,clip,width=\textwidth]{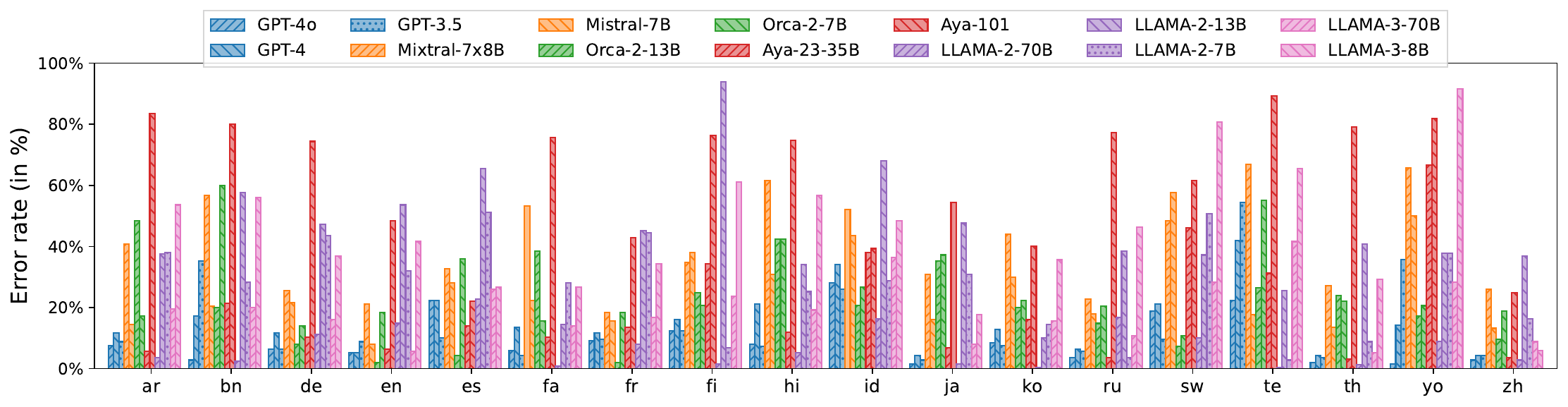}
        \caption{
        Error rate (in \%) = $\mathrm{FN}/(\mathrm{FN} + \mathrm{TP}$) on the relevant subset in \nomiracl test split. The relevant subset contains queries with known answers, i.e., at least one of the top-$k$ (where $k=10$) passages judged by a human annotator is relevant. On average, most LLMs (except Mistral and Aya-101) have a lower error rate, i.e., can accurately identify the relevant answer. Lower the error rate is better.}
        \label{fig:relevant-baseline-results-all}
    \end{center}
    \vspace{-\baselineskip}
\end{figure*}

All multilingual-focused LLM checkpoints used in our experiments for both closed and open-sourced can be found in \autoref{tab:model_links}. Hyperparameter choices during \nomiracl supervised fine-tuning LLMs are listed in \autoref{tab:hyperparameter_settings} and experimental results in \autoref{tab:fine-tuning-results-all}. LLM evaluation results for both the non-relevant and relevant subsets for all models can be found in \autoref{fig:non-relevant-baseline-results-all} and \autoref{fig:relevant-baseline-results-all} respectively. \autoref{fig:prompt_role_ablation} shows template changes for prompt optimization ablation experiments, including (i) role, (ii) repeat, and (iii) explanation prompts.


\begin{table}[tb!]
    \small
    \centering
    \resizebox{0.48\textwidth}{!}{\begin{tabular}{ l l }
        \toprule
        \multicolumn{1}{l}{\textbf{Model}} &
        \multicolumn{1}{c}{\textbf{Model Checkpoints (Link)}} \\ \midrule
        \multicolumn{2}{c}{\emph{OpenAI baseline models}} \\ \noalign{\vskip 0.5ex}\hdashline\noalign{\vskip 1ex}
        GPT-4o & \href{https://learn.microsoft.com/en-us/azure/ai-services/openai/}{learn.microsoft.com/en-us/azure/ai-services/openai/} \\
        GPT-4 & \href{https://learn.microsoft.com/en-us/azure/ai-services/openai/}{learn.microsoft.com/en-us/azure/ai-services/openai/} \\
        GPT-3.5 & \href{https://learn.microsoft.com/en-us/azure/ai-services/openai/}{learn.microsoft.com/en-us/azure/ai-services/openai/} \\ \midrule
        \multicolumn{2}{c}{\emph{Mistral baseline models}} \\ \noalign{\vskip 0.5ex}\hdashline\noalign{\vskip 1ex}
        Mixtral-8x7B & \href{https://huggingface.co/mistralai/Mixtral-8x7B-Instruct-v0.1}{huggingface.co/mistralai/Mixtral-8x7B-Instruct-v0.1} \\
        Mistral-7B (v0.3) & \href{https://huggingface.co/mistralai/Mistral-7B-Instruct-v0.3}{huggingface.co/mistralai/Mistral-7B-Instruct-v0.3} \\ \midrule
        \multicolumn{2}{c}{\emph{Orca-2 baseline models}} \\ \noalign{\vskip 0.5ex}\hdashline\noalign{\vskip 1ex}
        Orca-2-13B & \href{https://huggingface.co/microsoft/Orca-2-13b}{huggingface.co/microsoft/Orca-2-13b} \\
        Orca-2-7B & \href{https://huggingface.co/microsoft/Orca-2-7b}{huggingface.co/microsoft/Orca-2-7b} \\ \midrule
        \multicolumn{2}{c}{\emph{Aya baseline models}} \\ \noalign{\vskip 0.5ex}\hdashline\noalign{\vskip 1ex}
        Aya-101 & \href{https://huggingface.co/CohereForAI/aya-101}{huggingface.co/CohereForAI/aya-101} \\ 
        Aya-23-35B & \href{https://huggingface.co/CohereForAI/aya-23-35B}{huggingface.co/CohereForAI/aya-23-35B} \\ \midrule
        \multicolumn{2}{c}{\emph{LLAMA-2 baseline models}} \\ \noalign{\vskip 0.5ex}\hdashline\noalign{\vskip 1ex}
        LLAMA-2-70B & \href{https://huggingface.co/meta-llama/Llama-2-70b-chat-hf}{huggingface.co/meta-llama/Llama-2-70b-chat-hf} \\ 
        LLAMA-2-13B & \href{https://huggingface.co/meta-llama/Llama-2-13b-chat-hf}{huggingface.co/meta-llama/Llama-2-13b-chat-hf} \\
        LLAMA-2-7B &  \href{https://huggingface.co/meta-llama/Llama-2-7b-chat-hf}{huggingface.co/meta-llama/Llama-2-7b-chat-hf} \\
        \midrule
        \multicolumn{2}{c}{\emph{LLAMA-3 baseline models}} \\ \noalign{\vskip 0.5ex}\hdashline\noalign{\vskip 1ex}
        LLAMA-3 (70B) & \href{https://huggingface.co/meta-llama/Meta-Llama-3-70B-Instruct}{huggingface.co/meta-llama/Meta-Llama-3-70B-Instruct} \\
        LLAMA-3 (8B) & \href{https://huggingface.co/meta-llama/Meta-Llama-3-8B-Instruct}{huggingface.co/meta-llama/Meta-Llama-3-8B-Instruct} \\
        \bottomrule
    \end{tabular}}
    \caption{All models and checkpoint links used for \nomiracl evaluation.}
    \label{tab:model_links}
\end{table}

\begin{table}[tb!]
    \centering
    \small
     \resizebox{0.48\textwidth}{!}{\begin{tabular}{lc}
        \toprule
        \textbf{Hyperparameter} & \textbf{Value} \\
        \midrule
        \texttt{use\_peft} & \texttt{true} \\
        \texttt{torch\_dtype} & \texttt{bfloat16} \\
        \texttt{lora\_r} & \texttt{64} \\
        \texttt{lora\_alpha} & \texttt{16} \\
        \texttt{lora\_dropout} & \texttt{0.05} \\
        \texttt{lora\_target\_modules} & \multicolumn{1}{p{5cm}}{\texttt{\{q\_proj, k\_proj, v\_proj, o\_proj, gate\_proj, up\_proj, down\_proj\}}} \\
        \texttt{learning\_rate} & \texttt{3.0e-06} \\
        \texttt{lr\_scheduler\_type} & \texttt{cosine} \\
        \texttt{max\_seq\_length} & \texttt{4096} \\
        \bottomrule
    \end{tabular}}
    \caption{Hyperparameter settings chosen during LoRA supervised fine-tuning (SFT) Mistral-7B (v0.3) and LLAMA-3 (8B) instruct models on the \nomiracl development split.}
    \label{tab:hyperparameter_settings}
\end{table}

\begin{figure*}[t!]
\begin{mdframed}[backgroundcolor=gray!5]
    \small
    \texttt{\textcolor{blue}{You are an evaluator checking whether the question contains the answer within the provided contexts or not.} I will give you a question and several contexts containing information about the question. Read the contexts carefully. If any of the contexts answers the question, respond as either ``Yes, answer is present'' or ``I don't know'': \\\\
    QUESTION: \{query\} \\\\
    CONTEXTS:\\
    {[1]} \{Passage title\}: {\{Passage text\}}\\
    {[2]} \{Passage title\}: {\{Passage text\}}\\
    ... \\
    {[10]} \{Passage title\}: {\{Passage text\}}\\\\
    \textcolor{red}{Please remember to read all the contexts carefully. If any of the contexts answers the question: \{query\}, respond as either ``Yes, answer is present'' or ``I don’t know''.}\\\\
    OUTPUT:
}
\end{mdframed}

\begin{mdframed}[backgroundcolor=gray!5]
    \small
    \texttt{\textcolor{violet}{Read the query and the contexts carefully and provide a step-by-step explanation for your answer. If any of the contexts answers the question, respond as either ``Yes, answer is present'' or ``I don’t know''. You must strictly follow the output format with \#\# Reasoning: ... \#\# Answer: ``Yes, answer is present'' OR ``I don’t know''}.\\\\
    QUESTION: \{query\} \\\\
    CONTEXTS:\\
    {[1]} \{Passage title\}: {\{Passage text\}}\\
    {[2]} \{Passage title\}: {\{Passage text\}}\\
    ... \\
    {[10]} \{Passage title\}: {\{Passage text\}}\\\\
    OUTPUT: 
}
\end{mdframed}

\caption{All prompt ablations used in our experiments for LLM hallucination evaluation for all 18 languages in \nomiracl on both the relevant and non-relevant subsets. \textcolor{blue}{Role} prompt appends the role of the LLM at the beginning of the prompt (highlighted in \textcolor{blue}{blue}). \textcolor{red}{Repeat} prompt highlights the task by repeating instructions at the end of the prompt (highlighted in \textcolor{red}{red}). \textcolor{violet}{Explanation} prompt asks the model to provide a reasoning path and finally answer the question (highlighted in \textcolor{violet}{violet}).}
\label{fig:prompt_role_ablation}
\end{figure*}

\begin{table*}[t!]
    \centering
    \small
    \resizebox{\textwidth}{!}{\begin{tabular}{lccccccccccccccccccc}
        \toprule
        & \texttt{ar} & \texttt{bn} & \texttt{de} & \texttt{en} & \texttt{es} & \texttt{fa} & \texttt{fr} & \texttt{fi} & \texttt{hi} & \texttt{id} & \texttt{ja} & \texttt{ko} & \texttt{ru} & \texttt{sw} & \texttt{te} & \texttt{th} & \texttt{yo} & \texttt{zh} & \texttt{Avg.} \\
        \midrule
        \multicolumn{20}{l}{\emph{Hallucination Rates (in \%) on \nomiracl test split (non-relevant subset)}} \\ \midrule
        \textbf{Llama-3 (8B)} & 19.6 & 20.0 & 29.5 & 19.2 & 30.0 & 44.8 & 35.2 & 10.5 & 14.4 & 23.2 & 52.0 & 31.6 & 26.8 & 4.0 & 13.2 & 41.6 & 1.6 & 64.4 & 26.8 \\
        \textbf{Llama-3 (8B) (w/ SFT)} & 83.0 & 12.5 & 36.9 & 44.0 & 46.8 & 65.6 & 36.9 & 70.0 & 61.5 & 53.1 & 41.7 & 9.7 & 85.4 & 42.2 & 0.0 & 82.7 & 12.0 & 20.0 & 44.7 \\
        \textbf{Mistral-7B (v0.3)} & 40.4 & 63.2 & 38.2 & 42.8 & 17.2 & 52.4 & 47.6 & 16.1 & 39.6 & 30.8 & 44.4 & 28.8 & 41.6 & 14.8 & 74.0 & 58.8 & 23.6 & 45.2 & 40.0 \\
        \textbf{Mistral-7B (v0.3) (w/ SFT)} & 46.8 & 33.2 & 34.1 & 73.6 & 33.6 & 52.4 & 48.0 & 42.7 & 26.0 & 59.2 & 52.0 & 47.6 & 62.8 & 35.6 & 31.2 & 40.0 & 45.2 & 33.2 & 44.3 \\ \midrule
        \multicolumn{20}{l}{\emph{Error Rates (in \%) on \nomiracl test split (relevant subset)}} \\ \midrule
        \textbf{Llama-3 (8B)} & 53.6 & 56.0 & 36.8 & 41.6 & 26.8 & 26.8 & 34.4 & 61.2 & 56.8 & 48.4 & 17.6 & 35.6 & 46.4 & 80.8 & 65.6 & 29.2 & 91.7 & 6.0 & 45.3 \\
        \textbf{Llama-3 (8B) (w/ SFT)} & 6.0 & 70.8 & 32.4 & 31.2 & 24.0 & 14.6 & 21.2 & 8.8 & 14.9 & 40.6 & 24.0 & 77.2 & 8.3 & 32.5 & 87.5 & 4.2 & 75.5 & 50.4 & 34.7 \\
        \textbf{Mistral-7B (v0.3)} & 14.4 & 20.4 & 21.6 & 8.0 & 28.0 & 22.4 & 15.6 & 38.0 & 30.8 & 43.6 & 16.0 & 30.0 & 18.0 & 57.6 & 17.6 & 13.6 & 50.0 & 13.2 & 25.5 \\
        \textbf{Mistral-7B (v0.3) (w/ SFT)} & 46.4 & 60.8 & 46.0 & 12.4 & 46.8 & 36.4 & 35.6 & 42.0 & 69.6 & 37.2 & 37.6 & 42.0 & 25.2 & 69.6 & 66.4 & 56.8 & 52.0 & 46.8 & 46.1 \\
        
        \bottomrule
    \end{tabular}}
    \caption{Complete SFT results using the \nomiracl development dataset for LLAMA-3 (8B) and Mistral-7B (v0.3) LLMs across all languages in \nomiracl. Lower the hallucination and error rates (\%) is better.}
    \label{tab:fine-tuning-results-all}
\end{table*}

\newpage

\begin{table*}[t!]
    \resizebox{\textwidth}{!}{\begin{tabular}{l  c  c }
        \toprule
        \multicolumn{1}{c}{\textbf{Query}}   &
        \multicolumn{1}{c}{\textbf{Judged Passages}} & \multicolumn{1}{c}{\textbf{Relevance}} \\ \midrule
    \rowcolor{red!15} \multicolumn{1}{p{3cm}}{What is the name of the winner of the Abel Prize of 2022?} & \multicolumn{1}{p{19cm}}
   {[1] \textbf{Abel Prize}: The Abel Prize () is a Norwegian prize awarded annually by the King of Norway to one or more outstanding mathematicians. It is named after Norwegian mathematician Niels Henrik Abel (1802–1829) and directly modeled after the Nobel Prizes. It comes with a monetary award of 6 million Norwegian kroner (NOK) (€635,000 or \$740,000).} & 0 \\ \noalign{\vskip 1.5ex}
   \rowcolor{red!15} \multicolumn{1}{c}{} & \multicolumn{1}{p{19cm}}
   {[2] \textbf{MIT Department of Mathematics}: The current faculty of around 50 members includes Wolf Prize winner Michael Artin, Abel Prize winner Isadore Singer, and numerical analyst Gilbert Strang. } & 0 \\  \noalign{\vskip 1.5ex}
   \rowcolor{red!15} \multicolumn{1}{c}{} & \multicolumn{1}{p{19cm}}
   {[3] \textbf{Abel Prize}: After interest in the concept of the prize had risen in 2001, a working group was formed to develop a proposal, which was presented to the Prime Minister of Norway in May. In August 2001, the Norwegian government announced that the prize would be awarded beginning in 2002, the two-hundredth anniversary of Abel's birth. Atle Selberg received an honorary Abel Prize in 2002, but the first actual Abel Prize was awarded in 2003.} & 0 \\ \noalign{\vskip 1.5ex}
   \rowcolor{red!15} \multicolumn{1}{c}{} & \multicolumn{1}{p{19cm}}
   {[4] \textbf{Abel Prize}: The prize was first proposed in 1899, to be part of the celebration of the 100th anniversary of Niels Henrik Abel's birth in 1902. Shortly before his death in 1899, the Norwegian mathematician Sophus Lie proposed establishing an Abel Prize when he learned that Alfred Nobel's plans for annual prizes would not include a prize in mathematics. King Oscar II was willing to finance a mathematics prize in 1902, and the mathematicians Ludwig Sylow and Carl Størmer drew up statutes and rules for the proposed prize. However, Lie's influence waned after his death, and the dissolution of the union between Sweden and Norway in 1905 ended the first attempt to create an Abel Prize. } & 0 \\ \noalign{\vskip 1.5ex}
   \rowcolor{red!15} \multicolumn{1}{c}{} & \multicolumn{1}{p{19cm}}
   {[5] \textbf{Eötvös Loránd University}: Eötvös Loránd University (, ELTE) is a Hungarian public research university based in Budapest. Founded in 1635, ELTE is one of the largest and most prestigious public higher education institutions in Hungary. The 28,000 students at ELTE are organized into eight faculties, and into research institutes located throughout Budapest and on the scenic banks of the Danube. ELTE is affiliated with 5 Nobel laureates, as well as winners of the Wolf Prize, Fulkerson Prize and Abel Prize, the latest of which was Abel Prize winner Endre Szemerédi in 2012. } & 0 \\ \noalign{\vskip 1.5ex}
   \rowcolor{red!15} \multicolumn{1}{c}{} & \multicolumn{1}{p{19cm}}
   {[6] 
   \textbf{Abel Prize}: Anyone may submit a nomination for the Abel Prize, however, self-nominations are not permitted. The nominee must be alive; however, if the awardee dies after being declared as the winner, the prize will be awarded posthumously.} & 0 \\ \noalign{\vskip 1.5ex}
   \rowcolor{red!15} \multicolumn{1}{c}{} & \multicolumn{1}{p{19cm}}
   {[7] \textbf{Abel Prize}: The Norwegian Academy of Science and Letters declares the winner of the Abel Prize each March after recommendation by the Abel Committee, which consists of five leading mathematicians. Both Norwegians and non-Norwegians may serve on the Committee. They are elected by the Norwegian Academy of Science and Letters and nominated by the International Mathematical Union and the European Mathematical Society. The committee is of 2018 chaired by Norwegian mathematician Hans Munthe-Kaas (University of Bergen), and was before that, headed by Professor John Rognes.} & 0 \\ \noalign{\vskip 1.5ex}
   \rowcolor{red!15}  \multicolumn{1}{c}{} & \multicolumn{1}{p{19cm}}
   {[8] \textbf{Hans Munthe-Kaas}: Munthe-Kaas received Exxon Mobil Award for best PhD at NTNU, 1989, and the Carl-Erik Frōberg Prize in Numerical Mathematics 1996 for the paper \"Lie–Butcher theory for Runge–Kutta Methods\". Munthe-Kaas is elected member of the Norwegian Academy of Science and Letters, the Royal Norwegian Society of Sciences and Letters and the Norwegian Academy of Technological Sciences. Munthe-Kaas is the chair of the international Abel prize committee (2018-2022), he is President of the Scientific Council of Centre International de Mathématiques Pures et Appliquées (CIMPA) (2017–present) and he is Editor-in-Chief of Journal Foundations of Computational Mathematics (2017–present). Munthe-Kaas was secretary of Foundations of Computational Mathematics (2005–2011) and member of the Board of the Abel Prize in Mathematics (2010–2018). } & 0 \\ \noalign{\vskip 1.5ex}
   \rowcolor{red!15} \multicolumn{1}{c}{} & \multicolumn{1}{p{19cm}}
   {[9] \textbf{Science and technology in Hungary}: Among Hungary's numerous research universities, the Eötvös Loránd University, founded in 1635, is one of the largest and the most prestigious public higher education institutions in Hungary. The 28,000 students at ELTE are organized into eight faculties, and into research institutes located throughout Budapest. ELTE is affiliated with 5 Nobel laureates, as well as winners of the Wolf Prize, Fulkerson Prize and Abel Prize, the latest of which was Abel Prize winner Endre Szemerédi in 2012. } & 0 \\ \noalign{\vskip 1.5ex}
   \rowcolor{red!15} \multicolumn{1}{c}{} & \multicolumn{1}{p{19cm}}
   {[10] \textbf{Abel Prize}: The Abel Prize's history dates back to 1899, when its establishment was proposed by the Norwegian mathematician Sophus Lie when he learned that Alfred Nobel's plans for annual prizes would not include a prize in mathematics. In 1902 King Oscar II of Sweden and Norway indicated his willingness to finance a mathematics prize to complement the Nobel Prizes, but the establishment of the prize was prevented by the dissolution of the union between Norway and Sweden in 1905. It took almost a century before the prize was finally established by the Government of Norway in 2001, and it was specifically intended \"to give the mathematicians their own equivalent of a Nobel Prize.\" The laureates are selected by the Abel Committee, the members of which are appointed by the Norwegian Academy of Science and Letters. } & 0 \\ \noalign{\vskip 1.5ex}
   \bottomrule
    \end{tabular}}
    \caption{Randomly sampled example of a query on ``\emph{What is the name of the winner of the Abel Prize of 2022?}'' and top-10 judged passages in English (\texttt{en}) from the non-relevant subset (test split) in \nomiracl. Titles of each passage are marked in \textbf{bold}. The relevance judgment has been annotated manually by a native speaker.}
    \label{tab:examples_non_valid}
\end{table*}

\begin{table*}[t!]
    \resizebox{\textwidth}{!}{\begin{tabular}{l  c  c }
        \toprule
        \multicolumn{1}{c}{\textbf{Query}}   &
        \multicolumn{1}{c}{\textbf{Judged Passages}} & \multicolumn{1}{c}{\textbf{Relevance}} \\ \midrule
   \rowcolor{green!15} \multicolumn{1}{p{3cm}}{In which country Praia dos Pescadores is?} & \multicolumn{1}{p{19cm}}
   {[1] \textbf{Praia dos Pescadores (Albufeira)}: Praia dos Pescadores or the ``Fishermans Beach'' is a blue flag beach on the Atlantic south coast of the Algarve, in the district of Bairro dos Pescadores (Neighborhood of the Fisherman), Albufeira which is within the Municipality of Albufeira, Portugal. The beach is one of the two beaches which front the town of ``Albufeira'' with ``Praia do Túnel'' at the western end and ``Praia dos Pescadores"" lying to the eastern end of the towns seafront. The town and its beaches are located west by road of the regions capital of Faro. In the days before Albufeira had a harbour and mariner the ``Praia dos Pescadores'' was where all the local fishermen operated from and the beach scene would have been very different to the site you see today. Then the beach would have been full of brightly painted fishing boats pulled up on this beach when not at sea and much of the tourist activities took place on the ``Praia do Túnel''. Today the ``Praia dos Pescadores'' is now used for tourism and is a very busy beach especially in the summer season.} & 1 \\ \noalign{\vskip 1.5ex}
   \rowcolor{red!15}  \multicolumn{1}{c}{} & \multicolumn{1}{p{19cm}}
   {[2] \textbf{Praia do Túnel (Peneco)}: Praia do Túnel is a beach on the Atlantic south coast of the Algarve, in the town of Albufeira which is within the Municipality of Albufeira, Portugal. The beach is also known as ``Praia do Peneco'' and is one of the two beaches which front the town of ``Albufeira'' with ``Praia do Túnel'' at the western end and ``Praia dos Pescadores'' lying to the eastern end of the towns seafront. The town and its beaches are located west by road of the regions capital of Faro. The beach gets its name from a 20 meter long tunnel next to the tourist office in the middle of Albufeira which cuts through the cliffs linking the towns square to the beach. At the western end of the beach there is a promenade which ends at the cave known as the Xorino Grotto. According to 13th-century legend, the cave was used as shelter by the Moors after the Christian conquest of Albufeira. As well as the tunnel there are several other points of access to the beach including a lift, ramps and steps [...]} & 0 \\  \noalign{\vskip 1.5ex}
   \rowcolor{red!15} \multicolumn{1}{c}{} & \multicolumn{1}{p{19cm}}
   {[3] \textbf{Praia dos Pescadores (Albufeira)}: The beach is in length and is wide at low tide. The beach is divided by a protruding cliff from Praia do Túnel at the western end of the seafront. To the beaches eastern boundary is the Praia do Inatel and it is divided from that beach by a concrete pier which covers the outflow of the Ribeira de Albufeira (Albufeira River). There are also cliffs at the eastern end and to the back of the beach there is an amphitheatre of white houses of the district of Bairro dos Pescadores. The beach can also be accessed by an outdoor foot escalator from the Pau da Bandeira bluff located south of Bairro dos Pescadores down to the beach and Albufeira old town.} & 0 \\ \noalign{\vskip 1.5ex}
   \rowcolor{red!15} \multicolumn{1}{c}{} & \multicolumn{1}{p{19cm}}
   {[4] \textbf{Praia dos Pescadores (Albufeira)}: Praia dos Pescadores is an easily accessed beach with its large hard surface square at beach level. There are two car park's near-by, one of which, is at beach level, a short distance along the Avenida 25 de Abril within the old town. The second car park is at the top of the cliffs at Bairro dos Pescadores and is accessed via the outdoor escalator. To the back of the western end of the beach there a variety of restaurants many of which specialise in the local fish and seafood. The beach has several licensed concessions with opportunities to hire parasols and sun loungers. There are also many organised beach and water sport concessions from volleyball to boat trips and Parasailing. The beach also has toilet and shower facilities. During the summer months the beach is patrolled by lifeguards. In recent years the beach has been the focal point for the new year celebrations in the town. A temporary concert stage is erected on the Largo 25 de Abril and concerts are held to celebrate the new year. In the past the celebration has seen international bands appearing such as British reggae/pop band UB40 in 2009. The celebrations cumulate with a firework display held just of the beach on boats and pontoons just of the shoreline. } & 0 \\ \noalign{\vskip 1.5ex}
   \rowcolor{red!15} \multicolumn{1}{c}{} & \multicolumn{1}{p{19cm}}
   {[5] \textbf{Praia do Penedo}: Praia do Penedo is a beach within the Municipality of Aljezur, in the Algarve, Portugal. The beach is on the western Seaboard in the north west of the Algarve. The beach is south west of the village of Aljezur, and is north west, by road, from the regions capital of Faro. The beach of Praia do Penedo is inside the Vicentine Coast Natural Park, an area of outstanding natural beauty.} & 0 \\ \noalign{\vskip 1.5ex}
   \rowcolor{red!15} \multicolumn{1}{c}{} & \multicolumn{1}{p{19cm}}
   {[6] \textbf{Praia do Norte}: Praia do Norte is a civil parish of the municipality of Horta, located along the northern coast between Cedros and Capelo, on the Portuguese island of Faial, in the archipelago of the Azores. The population in 2011 was 250, in an area of . It is the least populous parish on the island, reached along the \"Estrada Regional\" E.R.1-1ª regional roadway from the urban centre of Horta.} & 0 \\ \noalign{\vskip 1.5ex}
   \rowcolor{red!15} \multicolumn{1}{c}{} & \multicolumn{1}{p{19cm}}
   {[7] \textbf{Praia das Conchas, São Tomé and Príncipe}: Praia das Conchas is a settlement in the western part of the Lobata District on São Tomé Island in São Tomé and Príncipe. Its population is 174 (2012 census). Established as a plantation (\"roça\"), Praia das Conchas lies 2 km from the coast, 3 km west of Guadalupe. There is a smaller seaside settlement also called \"Praia das Conchas\", 3.5 km to the north.} & 0 \\ \noalign{\vskip 1.5ex}
   \rowcolor{red!15} \multicolumn{1}{c}{} & \multicolumn{1}{p{19cm}}
   {[8] \textbf{Praia das Gatas}: Praia das Gatas (Portuguese meaning \"beach of the cats\") is a sandy beach in the northeastern part of the island of Boa Vista in Cape Verde. The nearest village is Fundo das Figueiras, 5km to the southwest. It forms a part of Northern Nature Park (\"Parque Natural do Norte\"). The small island Ilhéu dos Pássaros lies off the coast at the Praia das Gatas. } & 0 \\ \noalign{\vskip 1.5ex}
   \rowcolor{red!15} \multicolumn{1}{c}{} & \multicolumn{1}{p{19cm}}
   {[9] \textbf{Praia (Santa Cruz da Graciosa)}: Praia (officially São Mateus da Praia) is a Portuguese civil parish in the municipality of Santa Cruz da Graciosa, on the island of Graciosa, in the Azores. It still retains its former name locally, owing to the parish having once been the historical municipality of Praia. The population in 2011 was 836, in an area of 12.82 km².} & 0 \\ \noalign{\vskip 1.5ex}
\rowcolor{red!15}  \multicolumn{1}{c}{} & \multicolumn{1}{p{19cm}}
   {[10] \textbf{Praia Harbor}: Praia Harbor () is the port of the city of Praia in the southern part of the island of Santiago, Cape Verde. It is situated in a natural bay of the Atlantic Ocean. Since the latest modernization in 2014, it has 2 long quays, 3 shorter quays, a quay for fishing boats with fish processing installations, 2 container parks, 2 roll-on/roll-off ramps, and a passenger terminal. The total length of the quays is 863 m, and the maximum depth is 13.5 m. The port of Praia played an important role in the colonization of Africa and South America by the Portuguese. With 817,845 metric tonnes of cargo and 85,518 passengers handled (2017), it is the second busiest port of Cape Verde, after Porto Grande (Mindelo).} & 0 \\ \noalign{\vskip 1.5ex}
   \bottomrule
    \end{tabular}}
    \caption{Randomly sampled example of a query on ``\emph{In which country Praia dos Pescadores is?}'' and top-10 judged passages in English (\texttt{en}) from the relevant subset (test split) in \nomiracl. Titles of each passage are marked in \textbf{bold}. The relevance judgment has been annotated manually by a native speaker.}
    \label{tab:examples_valid}
\end{table*}

\end{document}